%% file: general_case__SSRN___arXiv_03-02-2026_.tex
\documentclass[11pt]{article}

\usepackage{fullpage}
\usepackage[utf8]{inputenc}
\usepackage{algorithm,algpseudocode}
\usepackage[T1]{fontenc}  
\usepackage{todonotes}
\usepackage{amsmath,amsfonts,amssymb,amsthm}
\usepackage{bbding,pifont}
\usepackage[margin=0.75in]{geometry}
\usepackage{url,hyperref}
\usepackage{xcolor}
\usepackage{float}
\usepackage{booktabs} 
\usepackage{graphicx}
\usepackage{caption}
\usepackage{subcaption}
\usepackage{ulem}
\usepackage{bm}
\usepackage[shortlabels]{enumitem}
\usepackage{tikz}
\usetikzlibrary{arrows.meta}

\allowdisplaybreaks
\input defs.tex

\newcommand{\norm}[1]{{\left\vert\kern-0.25ex\left\vert\kern-0.25ex\left\vert #1
    \right\vert\kern-0.25ex\right\vert\kern-0.25ex\right\vert}}
\numberwithin{equation}{section}

\title{Policy Transfer for Continuous-Time Reinforcement Learning: \\
A (Rough) Differential Equation Approach\footnote{This paper generalizes the theoretical results contained in an earlier and shorter version entitled ``Policy Transfer Ensures Fast Learning for Continuous-Time LQR with Entropy Regularization.''}}
\author{
Xin Guo
\thanks{UC Berkeley, IEOR. \textbf{Email:} \texttt{xinguo@berkeley.edu}}
\and 
Zijiu Lyu
\thanks{UC Berkeley, IEOR. \textbf{Email:} \texttt{zijiu.lyu@berkeley.edu}}
}
\date{}

\begin{document}

\maketitle

\begin{abstract}

    This paper studies policy transfer, one of the well-known transfer learning techniques adopted in large language models, for continuous-time reinforcement learning problems. In the case  of continuous-time linear-quadratic systems with Shannon's entropy regularization, we fully exploit the Gaussian structure of their optimal policy and the stability of their associated Riccati equations. In the general case where the system has possibly non-linear and bounded dynamics, the key technical component is the stability of diffusion SDEs which is established by invoking the rough path theory. Our work provides the first theoretical proof of policy transfer for continuous-time RL: an optimal policy learned for one RL problem can be used to initialize to search for a near-optimal policy for another closely related RL problem, while achieving (at least) the same rate of convergence for the original algorithm. As a byproduct of our analysis, we derive the stability of a concrete class of continuous-time score-based diffusion models via their connection with LQRs. 
    
    To illustrate the benefit of policy transfer for RL, we propose a novel policy learning algorithm for continuous-time LQRs, which achieves global linear convergence and local super-linear convergence. 
\end{abstract}

\section{Introduction}

\paragraph{Reinforcement learning (RL).}
Reinforcement learning (RL) is one of the fundamental machine learning paradigms, where an agent learns to make a sequence of decisions by interacting with an environment and possibly with other agents. In a typical RL setup, an agent learns a policy/strategy for choosing actions in a given system state through trial and error to maximize a cumulative reward over time. However, training an agent for a complex RL task from the ground up can be extremely inefficient.

\paragraph{Transfer learning (TL).}
Transfer learning is a machine learning technique that leverages expertise gained from one problem (called source task) to enhance the learning process in another related one (called target task). It is one of the most influential techniques that underpin the capabilities of large language models (LLMs). In the context of LLMs, transfer learning involves using pre-trained models, such as those from the GPT, BERT, or similar families, that were initially trained for specific tasks. Transfer learning repurposes these models for new and related applications, often involving domain-specific variations of the original problems. See \eg \cite{howard2018universal}, \cite{devlin2019bert}, \cite{raffel2020exploring}, \cite{brown2020language}, \cite{liu2019roberta}. Beyond LLMs, transfer learning has also gained a significant traction in other domains,  particularly for improving learning efficiency when data and computational resources are limited. See \eg \cite{kraus2017decision}, \cite{amodei2016deep}, \cite{tang2022self}.

\paragraph{RL with TL.}
Given the exponentially growing demand for complex RL tasks, and the increasing number of pre-trained RL models for various learning tasks, it is natural to incorporate TL into RL to leverage knowledge from a pre-trained RL model to reduce both training time and computational costs, especially when there is a limited amount of data for new RL models.

Policy transfer is one of the most direct methods to leverage knowledge from one RL task to another. The basic idea of policy transfer is to use the policy learned from the source task to initialize the policy for the target task. If two RL tasks are similar, exploring the pre-trained policy as a starting point hopefully allows the agent to find a near-optimal policy, with subsequent minor adjustments. This is intuitively clear and simple, and has been analyzed in a discrete-time LQ framework by \cite{guo2026fast}. Their work, as the first  known theoretical studies for incorporating TL into RL, demonstrates the algorithmic performance improvement with TL technique for RL.  

A natural question is whether the same benefit of transfer learning can be achieved for continuous-time RL via an appropriate policy transfer? Indeed, reinforcement learning, though primarily developed for discrete environment, is intrinsically continuous and complex, especially in robotics control, automatic driving, and portfolio optimization. However, analyzing transfer learning in the continuous-time RL framework remains uncharted and presents significantly greater technical challenges, as the knowledge to be transferred involves controlled stochastic processes and infinite-dimensional functional spaces.

\paragraph{Our work.}
This paper presents a theoretical analysis of policy transfer for  continuous-time RL problems. Our analysis starts with   a linear-quadratic system with a Shannon's entropy regularization term  (\emph{a.k.a.} LQRs). We show that an algorithm for the optimal policy of one LQR problem yields an optimal policy for any closely related LQR problem with the same convergence rate. 
The analysis for LQRs reveals  critical technical steps that we shall follow to analyze a general class of system which have possibly non-linear and bounded dynamics. In this general setting, we demonstrate that an optimal policy learned for one RL problem may serve as a near-optimal policy for any closely related RL problem, while ensuring the same convergence rate by the original learning algorithm. The main technical components of the analysis is the stability of diffusion SDEs,  established by the stability of associated \textit{rough differential equations} (RDEs) in the \textit{rough path theory}. 

To illustrate explicitly the benefit of policy transfer for RL, we  propose a novel policy learning algorithm for continuous-time LQRs, which achieves global linear convergence and local super-linear convergence. This implies that any closely related LQR is guaranteed with a super-linear convergent learning algorithm. Our analysis fully exploits the Gaussian structure of the optimal policy for LQRs, as well as the robustness of the associated Riccati equation. 

As a byproduct of our analysis, we derive the stability of a class of continuous-time score-based diffusion models via their connection with LQRs. The key ingredient is the Cole-Hopf transformation, which transforms the HJB equation of the LQRs into the Fokker-Planck equation of an O-U process. 

Our results demonstrate both theoretical guarantees and algorithmic benefits of transfer learning in continuous-time RL, filling a gap in the existing literature and expanding prior work from discrete to continuous time settings.

\paragraph{Related work.}
The existing literature on policy learning for Linear-Quadratic (LQ) problems is extensive. Notable examples focusing on gradient-based algorithms for discrete-time LQRs include \cite{fazel2018global} and \cite{hambly2021policy}, which achieve global linear convergence in learning the parameters of the optimal feedback policy. \cite{giegrich2024convergence} extends this approach to continuous-time LQRs, also demonstrating global linear convergence. Beyond these gradient-based methods, other aspects of LQRs have also been explored. For instance, \cite{dean2020sample} develops a multistage procedure for designing a robust controller of discrete-time LQRs when the system dynamics are not fully known, while \cite{basei2022logarithmic} (\resp \cite{huang2024sublinear}) introduces an algorithm that is theoretically guaranteed with a logarithmic (\resp sublinear) regret bound. Furthermore, \cite{krauth2019finite} provides a theoretical analysis of the sample complexity of approximate policy iteration for learning discrete-time LQRs. For a more comprehensive background, interested readers are referred to the standard references by \cite{kwakernaak1972linear} and \cite{bertsekas2019reinforcement}.

The stability of diffusion SDEs and stochastic control problems has also been studied  extensively: for instance,  \cite[proof of Theorem~5.2.1, pp.71-72]{SDE} shows that the marginal distribution of the solution of an It\^o diffusion SDE is $L^2$-stable with respect to the SDE's initial condition, \cite[Theorem~5.3]{khasminskii2012stochastic} proves that the solution of an It\^o diffusion SDE is \textit{stable in probability} by generalizing the Lyapunov theorem for deterministic systems, the Wong-Zakai theorem (\cf \cite{wong1965convergence, eugene1965relation}) shows diffusion SDE's stability with respect to the driving path, and the rough path theory (see \eg \cite{friz2010multidimensional, friz2014course, lyons2007differential}) establishes the stability of RDEs which can be viewed as a generalization of diffusion SDEs. For the latter, \cite{pradhan2023robustness} shows the stability of a stochastic control problem with respect to the system's driving path by RDEs' stability, and \cite{pradhan2024robustness} proves that the value function of a stochastic control problem is robust with respect to the diffusion SDE's dynamics by exploiting the strong regularity property of HJB equation's solution.

Our work of RL with TL, especially analysis for the LQRs, is inspired by \cite{guo2026fast}, where a super-linear local convergent algorithm called IPO is proposed for discrete-time exploratory LQRs. In comparison, our analysis of policy transfer between continuous-time LQRs is technically more challenging. More importantly, we establish general results on policy transfer between any two closely related RL problems. The particular IPO algorithm for LQRs illustrates the benefit of policy transfer in such a context. 

On the stability of continuous-time score-based diffusion models, \cite{tang2025score} has obtained a fairly general result, assuming appropriate technical conditions. Here we present, via connecting score-based diffusion models with LQRs, a concrete class of models where these assumptions are explicitly verified to ensure the stability results.
Finally, the connection between score-based diffusion models and LQRs is well known. For example, \cite{zhang2023mean} shows that a large class of generative models, including normalizing flows, score-based diffusion models, and Wasserstein gradient flows, can be viewed as the solutions to certain mean-field games (MFGs). We note that LQRs can be viewed as the degenerate case of LQMFGs. Moreover, \cite{gu2024combining} and \cite{zhang2024wasserstein} analyze the relationship between MFGs, Wasserstein proximals and score-based diffusion models.

\paragraph{Organization of the paper.}
The reminder of the paper is organized as follows. In Section~\ref{sec: LQR}, we formulate the LQRs mathematically and establish the transfer learning between two LQRs via the stability of the associated Riccati equations. Then, in Section~\ref{sec: general}, we consider the second class of RL problems where the system has possibly non-linear and bounded dynamics, and present the corresponding transfer learning results. Next, in Section~\ref{sec: LQR-transfer}, we derive the IPO algorithm for the optimal policy of LQRs and show its global linear and local super-linear convergence. In Section~\ref{sec: application}, we show the stability of a class of score-based diffusion models via their connection with LQRs. Finally, the proof of all lemmas and propositions are given in Section~\ref{sec: proofs}, with the appendix for a gentle introduction of the rough path theory.

\paragraph{Notation.}
For any smooth function $f: \R^n \rightarrow \R$, we use $\nabla f(x) \in \R^n$ to denote its gradient, and $\Delta f(x) \in \R^{n \times n}$ to denote its Hessian matrix. In addition, we use $\cdot$ to indicate the usual vector-vector and matrix-matrix inner products, depending on the context, and we use $S^n_{\geq 0}$ (\resp $S^n_{> 0}$) to denote the space of $n \times n$ real positive semi-definite (\resp positive definite) matrices.

\section{Transfer learning between RL problems}

\subsection{Transfer learning between LQRs with entropy regularization} \label{sec: LQR}

We start with the study of transfer learning between entropy-regularized continuous-time linear quadratic regulators (LQRs). This analysis will provide important insight for a more general class of RL problems. 

To start, let us fix the mathematical framework under which LQRs are defined over a finite time interval $[0, T]$. Specifically, we  assume that the state process $x_t \in \R^n$ of the agent  follows the linear SDE:
\begin{equation} \label{eq: state}
    \dd x_t = \Big[A_t x_t + B_t \E(u_t \,|\, x_t) \Big] \dd t + \sigma_t \dd W_t, \quad x_0 \sim \D_0,
\end{equation}
where $\E[u_t \,|\, x_t] \sim \pi_t(\cdot \,|\, x_t) \in \pP(\R^k)$ represents the randomized policy of the agent conditioned on $x_t$, $(W_t)_{t \in [0, T]}$ denotes the $d$-dimensional standard Brownian motion ($d$-BM for short), $\D_0$ denotes the initial distribution, and $(A_t, B_t, \sigma_t)_{t \in [0, T]}$ are appropriate deterministic matrix-valued processes to be specified later. 
The agent aims to minimize the following entropy-regularized cost function:
\begin{equation} \label{eq: cost}
    \inf_{\pi \in \A} J_\pi (0, \D_0) := \E_{u_t \sim \pi_t(\cdot \,|\, x_t)}\Bigg[ \int_0^T x_t^\dagger Q_t x_t + u_t^\dagger R_t u_t + \tau \log h_t (u_t \,|\, x_t) \dd t + x_T^\dagger Q^\prime x_T \,\Bigg|\, x_0 \sim \D_0 \Bigg],
\end{equation}
where $^\dagger$ denotes the transpose operator, $\A$ denotes the set of admissible randomized policies, $h_t (\cdot \,|\, x_t)$ denotes the conditional probability distribution function of $\pi_t(\cdot \,|\, x_t)$ (\ie Shannon's entropy\footnote{We note that our analysis of transfer learning can be generalized to other entropies, \eg cross entropy.}), and $(Q_t, R_t)_{t \in [0, T]}$ (\resp $Q^\prime$) are appropriate deterministic matrix-valued processes (\resp matrix) to be specified later.
Note that the exploratory SDEs adopted here are first proposed by \cite{wang2018exploration}, where an entropy-regularization term is added to the cost function to encourage agent exploration.

Next, we present the technical assumptions to ensure that the above formulation \eqref{eq: state} -- \eqref{eq: cost} is well-defined. In particular, our goal is to ensure that \eqref{eq: state} admits a unique strong solution (see \eg \cite[Theorem 5.2.1]{SDE}) and that \eqref{eq: cost} has a finite integrand. See \cite[Section~2]{guo2022entropy} for a similar setup.

\begin{assumption}[Probability space] \label{ass: prob}
    The filtered probability space $\left(\Omega, \F, \PP; (\F_t)_{t \in [0, T]}\right)$ with the filtration $(\F_t)_{t \in [0, T]}$ 1) is rich enough to support some $d$-BM $(W_t)_{t \in [0, T]}$, the random action $(u_t)_{t \in [0, T]}$ of the agent, and the initial distribution $\D_0$, which are assumed to be independent; 2) satisfies the usual conditions (\ie $\F_0$ contains all the $\PP$-null sets and $(\F_t)_{t \in [0, T]}$ is right-continuous).
\end{assumption}

\begin{assumption}[Admissible policies] \label{ass: policy}
    The set $\A$ of admissible policies consists of Markovian randomized policies, \ie the following conditions hold for any $\pi \in \A$:
    \begin{enumerate}[1)]
        \item for any $t \in [0, T]$ and $x \in \R^n$, $\pi_t(\cdot \,|\, x)$ is absolutely continuous w.r.t. the Lebesgue measure on $\R^k$ and has a finite expectation and a finite entropy;
        \item $\E(u_t \,|\, x)$, when viewed as a function of $(t, x) \in [0, T] \times \R^n$, has a linear growth w.r.t. $x$ and is Lipchitz continuous in $x$.
    \end{enumerate}
\end{assumption}

\begin{assumption}[Regularity conditions] \label{ass: regular}
    $\D_0$ is assumed to be square integrable, and
    \begin{equation*}
        \begin{gathered}
            A, Q \in L^\infty([0, T], \R^{n \times n}), \quad B \in L^\infty([0, T], \R^{n \times k}), \\
            R \in L^\infty([0, T], \R^{k \times k}), \quad \sigma \in L^2([0, T], \R^{n \times d}).
        \end{gathered}
    \end{equation*}
 $Q_t \succeq 0$ a.e. for $t \in [0, T]$, $\tau > 0$, $Q^\prime \succeq 0$, and there exists $\delta > 0$ such that $R_t - \delta I \succeq 0$ a.e. for $t \in [0, T]$.
\end{assumption}

Now, we consider the transfer learning between two LQRs, whose system parameters are $(\theta_t)_{t \in [0, T]}$ and $(\tilde{\theta}_t)_{t \in [0, T]}$, respectively. Without loss of generality,  assume that the first LQR is accessible, with its optimal policy given  by the parameter of $(K^\ast_t)_{t \in[0, T]}$. We will show that if $(\theta_t)_{t \in [0, T]}$ and $(\tilde{\theta}_t)_{t \in [0, T]}$ are sufficiently close, then $(K^\ast_t)_{t \in[0, T]}$ may be used as an initialization to efficiently learn the optimal policy of the second LQR. Here in our framework the parameters $\theta = (A, Q, B, R, Q^\prime, \tau)$. Since the optimal policy does not depend on the diffusion coefficient $(\sigma_t)_{t \in [0, T]}$ and the initial distribution $\D_0$ (\cf Lemma~\ref{lemma:optimalpolicy}), we exclude these two parameters from our subsequent discussion. 

\begin{theorem}
    [Transfer learning of LQRs] \label{thm: general_TL}
    Suppose there is an LQR whose model parameters are $\theta$. Let $\{\pi^{(i)}_\theta\}_{i \geq 0}$ be a sequence of Gaussian policies that converges to an optimal (Gaussian) policy of the LQR. Then, for any $\epsilon > 0$, there exists $\zeta > 0$ and $N > 0$, such that each policy in $\{\pi^{(i)}_\theta\}_{i \geq N}$ is $\epsilon$-optimal in the LQR whose model parameters are $\tilde{\theta}$ such that $d(\tilde{\theta}, \theta) < \zeta$. Here $d$ denotes an appropriately chosen distance on the metric space of Gaussian policies.
\end{theorem}

This result is based on the following two lemmas. First, we see that the optimal randomized policy of the LQR defined by \eqref{eq: state} -- \eqref{eq: cost} can be derived via the dynamic programming principle (DPP) and by following a similar calculation from the earlier work \cite{wang2018exploration} and \cite{guo2022entropy}. 

\begin{lemma}\label{lemma:optimalpolicy}
    The optimal randomized policy of the LQR \eqref{eq: state} -- \eqref{eq: cost} is:
    \begin{equation}
        \pi^\ast_t (\cdot \,|\, x) = \mathcal{N} \left(- R_t^{-1} B_t^\dagger P_t x,\, \frac{\tau}{2} R_t^{-1} \right), \label{eq: NE_policy}
    \end{equation}
    where $P_t$ solves the following Riccati equation:
    \begin{equation}
        \frac{\dd P_t}{\dd t} + A_t^\dagger P_t + P_t A_t + Q_t - P_t B_t R^{-1}_t B_t^\dagger P_t = 0, \quad P_T = Q^\prime. \label{eq: P_t_1}
    \end{equation}
\end{lemma} 

\begin{remark}
    The Gaussian form of $\pi^\ast$ originates from the entropy-regularization term in the cost function \eqref{eq: cost}. The mean of $\pi^\ast$ appears in a mean-reverting fashion, pushing the agent to 0. Meanwhile, the covariance of $\pi^\ast$ is driven by the regularization coefficient $\tau > 0$. The larger the value of $\tau$, the more the agent would explore. In the case where $\tau \rightarrow 0^+$, $\pi^\ast$ would converge to a deterministic policy as one should expect (see \cite[Section~5.4]{wang2018exploration}). 
\end{remark}

\begin{lemma}[Stability of Riccati equation] \label{lemma: R_continuity}
    Under Assumption~\ref{ass: regular}, denote by $\mathcal{R}$ the solution map of the Riccati equation \eqref{eq: P_t_1}, \ie
    \begin{align*}
        \mathcal{R} : L^\infty([0, T], \R^{n \times n}) \times L^\infty([0, T], S^n_{\geq 0}) & \times L^\infty([0, T], \R^{n \times k}) \times L^\infty([0, T], S^k_{> 0}) \times S^n_{\geq 0} \\
        & \longrightarrow C([0, T], S^n_{\geq 0}) \\
        (A_{t \in [0, T]}, Q_{t \in [0, T]}, B_{t \in [0, T]}, R_{t \in [0, T]}, Q^\prime) & \longmapsto \mathcal{R}(A, Q, B, R, Q^\prime) := (P_t)_{t \in [0, T]}.
    \end{align*}
    Then, $\mathcal{R}$ is continuous, with the $L^\infty$ (\resp $S^n_{\geq 0}$) space  equipped with the uniform norm $||\cdot||_{\infty; [0, T]}$ (\resp matrix 2-norm).
\end{lemma}
Then, Theorem \ref{thm: general_TL} follows immediately from Lemma~\ref{lemma: R_continuity} (\cite{basei2022logarithmic}). Indeed, by Lemma \ref{lemma: R_continuity},  the optimal policy is a continuous function in the LQR's model parameters $\theta$. As a result, when the distance between $\tilde{\theta}$ and $\theta$ is small enough, the optimal policies of the two LQRs can be made arbitrarily close to each other. This implies the desired near-optimality.

\subsection{Transfer learning between general continuous-time RL problems} \label{sec: general}

In this section, we consider transfer learning between continuous-time RL problems beyond the LQR framework. As seen from the analysis for the LQRs, it is crucial to establish the stability of the associated optimal policy. Unlike LQRs where the optimal policy has a simple Gaussian structure and   such stability can be established through the stability of the associated Riccati equation, we instead rely on the stability of associated RDEs. 

Let us start with  the following  stochastic  control problem, where we assume the  state to follow the time-homogeneous Stratonovich SDE:
\begin{equation} \label{eq: Strat_SDE}
    \dd X^\alpha_t = \mu(X^\alpha_t, \alpha_t) \dd t + \sigma(X^\alpha_t) \circ \dd W_t,
\end{equation}
where $W_{[0, T]}$ is a $d$-BM, $\alpha_t := \alpha(X_t) \in \R^k$ denotes the control process (\ie we only consider Markov feedback controls), $X_t^\alpha$ denotes the controlled process (\ie we use the superscript $^\alpha$ to imply the dependence), and $\mu : \R^n \times \R^k \rightarrow \R^n$ and $\sigma : \R^n \rightarrow \R^{n \times d}$ are deterministic functions which satisfy certain regularity conditions. We denote by $J$ the cost function such that
\begin{equation} \label{eq: general_cost}
    J: \mathcal{P}\big(C([0, T], \R^n)\big) \times \A \longrightarrow \R,
\end{equation}
which satisfies certain regularity conditions. Here $C([0, T], \R^n)$ denotes the space of continuous functions on $[0, T]$, equipped with the uniform norm $||\cdot||_{\infty; [0, T]}$, $\mathcal{P}\big(C([0, T], \R^n)\big)$ denotes the space of Borel probability measures on $C([0, T], \R^n)$, equipped with the weak topology, and $\A$ denotes the space of admissible (Markov feedback) controls. We view the solution of \eqref{eq: Strat_SDE} (\ie the law of $X^\alpha_{[0, T]}$, which is denoted by $\mathcal{L}(X^\alpha_{[0, T]})$) as an element in $\mathcal{P}\big(C([0, T], \R^n)\big)$.

\begin{remark}
    By treating time $t$ as a coordinate of the state, one can always transform a time-inhomogeneous SDE into a time-homogeneous one. Therefore, the time-homogeneity of \eqref{eq: Strat_SDE} is assumed without loss of generality. In addition, we formulate \eqref{eq: Strat_SDE} in the sense of Stratonovich because of its connection with the \textrm{rough path theory}, which is our main technical tool in this section. In fact, under certain regularity conditions, there is a one-to-one correspondence between Stratonovich SDEs and It\^o SDEs, see \eg \cite[Chapter~6]{evans2012introduction}. 
\end{remark}

We impose the following technical assumptions on the control problem given by \eqref{eq: Strat_SDE} -- \eqref{eq: general_cost} to ensure its well-posedness.

\begin{assumption}[Regularity conditions] \label{ass: general_regularity}
    The initial condition of \eqref{eq: Strat_SDE} is given by $X_0 \in L^2 (\Omega, \mathcal{F}, \mathbb{P}; \R^n)$ for some given Polish atomless probability space which is independent of $W_{[0, T]}$; the function $\mu \in {\rm Lip}^1 (\R^n \times \R^k, \R^n)$, \ie $\mu$ is bounded and Lipschitz continuous; the function $\sigma \in {\rm Lip}^2 (\R^n, \R^{n \times d})$, \ie $\sigma$ is bounded, and has a bounded and Lipschitz continuous first-order derivative; and the cost function $J$ is continuous (we will specify $\A$ in Assumption~\ref{ass: general_admissible}).
\end{assumption}

\begin{remark}
    Our formulation and Assumption~\ref{ass: general_regularity} of general control problems are subtly different from the classical stochastic control setting (see \eg \cite[Section~3.2]{pham2009continuous}) in the following aspects:
    \begin{itemize}
        \item The coefficients $\mu$ and $\sigma$ in \eqref{eq: Strat_SDE} are required to be uniformly bounded, instead of Lipschitz continuous w.r.t. $x$ in the classical setting (\eg O-U processes do not fit into the framework of Section~\ref{sec: general});
        \item The cost function $J$ in \eqref{eq: general_cost} is required to be a continuous function on $\mathcal{P}\big(C([0, T], \R^n)\big) \times \A$. By the definition of weak convergence, a class of choices for the cost function $J$ that satisfies Assumption~\ref{ass: general_regularity} is the integral of a bounded continuous function defined on $C([0, T], \R^n) \times \A$. In the classical setting, the cost function is commonly written as the expectation of a measurable function in $(t, x, a)$.
    \end{itemize}
\end{remark}

We always equip ${\rm Lip}^1$ with the topology induced by uniform convergence on compact sets. That is, $\mu_n \rightarrow \mu$ in ${\rm Lip}^1$ if and only if $||\mu_n - \mu||_{\infty; \mathcal{K}} \rightarrow 0$ for every compact subset $\mathcal{K}$. Similarly, we equip ${\rm Lip}^2$ with the topology such that $\sigma_n \rightarrow \sigma$ in ${\rm Lip}^2$ if and only if $||\sigma_n - \sigma||_{\rm Lip^1; \mathcal{K}} \rightarrow 0$ for every compact subset $\mathcal{K}$, where
\[
||f||_{\rm Lip^1; \mathcal{K}} := \max\Big\{ ||f||_{\infty; \mathcal{K}},\, ||\nabla f||_{\infty; \mathcal{K}},\, ||\nabla f||_{\text{1-Hol}; \mathcal{K}} \Big\}
\]
and $||\cdot||_{\text{1-Hol}}$ denotes the 1-Holder norm (\cf \cite[Definition 5.1(i)]{friz2010multidimensional}). We note that both ${\rm Lip}^1$ and ${\rm Lip}^2$ are metrizable. We use $||\cdot||_{\infty;\text{compact}}$ and $||\cdot||_{{\rm Lip}^1;\text{compact}}$ to denote their corresponding metrics. 

\begin{assumption}[Admissible set of controls] \label{ass: general_admissible}
    The admissible set $\mathcal{A}$ of Markov feedback controls is a compact subspace of ${\rm Lip} (\R^n, \R^k)$, which denotes the space of Lipschitz continuous functions equipped with the metric $||\cdot||_{\infty;\rm{compact}}$. 
\end{assumption}

\begin{remark}
    Examples of $\mathcal{A}$ can be constructed utilizing the Arzela-Ascoli theorem (\ie uniform boundedness combined with  equicontinuity implies  uniform convergence on compact sets).
\end{remark}

Studies of transfer learning between two general continuous-time RL problems rely on the continuity of the map
\newpage
\begin{align*}
    F: \mathcal{A} & \longrightarrow \R, \\
    \alpha & \longmapsto J \big(\mathcal{L}(X^\alpha_{[0, T]}), \alpha\big),
\end{align*}
where $\mathcal{L}(X^\alpha_{[0, T]})$ denotes the law of the controlled state process. The continuity of $F$ is built on two lemmas. 

The first lemma states that Stratonovich diffusion SDEs can be solved as \textit{rough differential equations} (RDEs) (\cf \cite[Theorem~17.3]{friz2010multidimensional}) from the \textit{rough path theory} (see \eg \cite{friz2010multidimensional, friz2014course, lyons2007differential} and also the appendix for an introduction).

\begin{lemma} \label{lemma: RDE}
    Let $\mu \in {\rm Lip}^1 (\R^n, \R^n)$, $\sigma \in {\rm Lip}^2 (\R^n, \R^{n \times d})$, and $y_0 \in \R^n$, with definitions of ${\rm Lip}^1$ and ${\rm Lip}^2$ maps given from Assumption~\ref{ass: general_regularity}. Then, the unique strong solution of the Stratonovich diffusion SDE
    \begin{equation*}
        \dd Y_t = \mu(Y_t) \dd t + \sigma(Y_t) \circ \dd W_t, \quad Y_0 = y_0
    \end{equation*}
    is indistinguishable to the (almost surely well-defined and unique) solution of the RDE
    \begin{equation} \label{eq: RDE_Strat}
        \dd Y_t = \mu(Y_t) \dd t + \sigma(Y_t) \dd \mathbf{W_t}, \quad Y_0 = y_0,
    \end{equation}
    where $\mathbf{W}_t$ is the enhanced BM in the sense of Stratonovich (\cf Example~\ref{exam: BM} in the appendix).
\end{lemma}

The second lemma utilizes the stability of RDEs to establish the continuity of the solution map of Stratonovich diffusion SDEs. 

\begin{lemma} \label{lemma: SDE_stability}
    Let $\mu \in {\rm Lip}^1 (\R^n, \R^n)$, $\sigma \in {\rm Lip}^2 (\R^n, \R^{n \times d})$ and $Y_0 \in L^2(\Omega, \F, \mathbb{P}; \R^n)$ which is independent to $W_{[0, T]}$. Denote by $\mathcal{R}$ the solution map of the following Stratonovich diffusion SDE:
    \begin{equation*}
        \dd Y_t = \mu(Y_t) \dd t + \sigma(Y_t) \circ \dd W_t,
    \end{equation*}
    that is,
    \begin{align*}
        \mathcal{R}: {\rm Lip}^1 (\R^n, \R^n) \times {\rm Lip}^2 (\R^n, \R^{n \times d}) \times L^2(\Omega, \F, \mathbb{P}; \R^n) & \longrightarrow \mathcal{P}\big(C([0, T], \R^n)\big), \\
        (\mu, \sigma, Y_0) & \longmapsto \mathcal{L}(Y_{[0, T]}).
    \end{align*}
    Then, $\mathcal{R}$ is continuous, where ${\rm Lip}^1$ (\resp ${\rm Lip}^2$, $L^2$) is equipped with the $||\cdot||_{\infty;\text{compact}}$ (\resp $||\cdot||_{{\rm Lip}^1;\text{compact}}$, $||\cdot||_{L^2}$) metric, and $\mathcal{P}\big(C([0, T], \R^n)\big)$ is equipped with the weak topology.
\end{lemma}

Now the continuity of $F$ follows immediately from Lemma~\ref{lemma: SDE_stability}. Moreover, given the compactness of $\mathcal{A}$ in Assumption~\ref{ass: general_admissible}, 
 the well-posedness of the control problem \eqref{eq: Strat_SDE} -- \eqref{eq: general_cost} is established. That is,    

\begin{proposition}[Well-posedness] \label{prop: existence}
    Under Assumptions~\ref{ass: general_regularity} -- \ref{ass: general_admissible}, the control problem \eqref{eq: Strat_SDE} -- \eqref{eq: general_cost} admits an optimal Markov feedback control in $\A$.
\end{proposition}

In fact, the above two lemmas naturally lead to the following result (\cf Theorem \ref{thm: general_transfer}) regarding the transfer learning between two continuous-time RL problems. To see this, consider the class of control problems which are in the same form as \eqref{eq: Strat_SDE} -- \eqref{eq: general_cost} and satisfy Assumptions~\ref{ass: general_regularity} -- \ref{ass: general_admissible}. Assume that these control problems  share the same admissible set $\mathcal{A}$ of Markov feedback controls as specified in Assumption \ref{ass: general_admissible} and cost function $J$. Now, let us view $\theta := (\mu, \sigma, X_0) \in {\rm Lip}^1 \times {\rm Lip}^2 \times L^2$ as the model parameters. For each $\theta$, we use $\mathfrak{C}(\theta)$ to denote the corresponding general control problem. Then by Lemma~\ref{lemma: SDE_stability},   the  map
\begin{align*}
    \mathcal{M} : \mathcal{A} \times {\rm Lip}^1 \times {\rm Lip}^2 \times L^2 & \longrightarrow \R, \\
    (\alpha, \mu, \sigma, X_0) & \longmapsto J \big(\mathcal{L}(X^\alpha_{[0, T]}), \alpha\big)
\end{align*}
is continuous, where $X^\alpha_{[0, T]}$ is the controlled process in \eqref{eq: Strat_SDE}. Therefore, the restriction of $\mathcal{M}$ to any compact subspace is uniformly continuous. That is, when the distance between $\theta$ and $\tilde{\theta}$ is sufficiently small, $\alpha^\ast$ becomes a near-optimal Markov feedback control of $\mathfrak{C}(\tilde{\theta})$. 

\begin{theorem}[Transfer learning] \label{thm: general_transfer}
    Let $\mathcal{B}$ be a compact subspace of ${\rm Lip}^1 \times {\rm Lip}^2 \times L^2$ and $\theta \in \mathcal{B}$. Suppose $\{\pi^{(i)}_\theta\}_{i \geq 0} \subset \mathcal{A}$ is a sequence of Markov feedback controls that converges to an optimal Markov feedback control of $\mathfrak{C}(\theta)$. Then, for any $\epsilon > 0$, there exists $\zeta > 0$ and $N > 0$, such that each control in $\{\pi^{(i)}_\theta\}_{i \geq N}$ is $\epsilon$-optimal in $\mathfrak{C}(\tilde\theta)$ where $\tilde\theta \in \mathcal{B}$ and $d(\theta, \tilde\theta) < \zeta$. Here $d$ denotes the corresponding metric on the metric spaces.
\end{theorem}

\begin{remark}
    The stability of diffusion SDEs (\cf Lemma~\ref{lemma: SDE_stability}) may be obtained (to some extent) by the Girsanov theorem (see \eg \cite[Section~8.6]{SDE}) as well, under a new set of regularity conditions on $\mu$ and $\sigma$ (\eg $\sigma$ is uniformly bounded away from 0). A naive application of the Girsanov theorem shows the weak convergence of marginal distributions, which is, however, strictly weaker than the weak convergence on the path space $C([0, T], \R^n)$ (\cf \cite[Example~2.7]{billingsley2013convergence}). 
\end{remark}

\begin{remark}
    Lemma~\ref{lemma: SDE_stability} is different from the Wong-Zakai theorem (\cf \cite{wong1965convergence, eugene1965relation}, also see \eg \cite[Theorem~9.3]{friz2014course}) in the following aspects. First, the Wong-Zakai theorem mainly considers the continuity of the Stratonovich SDE's solution w.r.t. the driving path, while Lemma~\ref{lemma: SDE_stability} considers the continuity w.r.t. the vector field and initial condition. Second, the Wong-Zakai theorem shows the almost surely convergence under the $\alpha$-Holder norm ($\alpha < \frac{1}{2}$), while Lemma~\ref{lemma: SDE_stability} only shows the weak convergence (in fact, convergence in probability) and the path space is equipped with the uniform norm which is strictly weaker than the $\alpha$-Holder norm. 
\end{remark}

\begin{remark}
    In \eqref{eq: Strat_SDE}, the diffusion coefficient $\sigma$ is not controlled. However, our framework can  be modified to make $\sigma$ also a controlled process. In that case, one needs more considerations when defining the admissible set $\mathcal{A}$ of Markov feedback controls, and the topology on $\mathcal{A}$ also needs to be refined to guarantee the well-posedness of the modified problem. In the classical setting, the counterpart of Proposition~\ref{prop: existence} commonly requires the diffusion coefficient $\sigma$ to satisfy the uniform ellipticity condition in order to invoke the PDE theory (\cf \cite[Remark~3.5.6]{pham2009continuous}), which is not required in our case.
\end{remark}

\begin{remark}
From the analysis in the previous section, it is clear that the assumption of $J$ can be further relaxed to ensure the results hold in Lemma \ref{lemma: SDE_stability}, Proposition \ref{prop: existence}, and Theorem \ref{thm: general_transfer}, as long as the following map is continuous:
    \begin{align*}
        \mathcal{M} : \mathcal{A} \times {\rm Lip}^1 \times {\rm Lip}^2 \times L^2 & \longrightarrow \R, \\
        (\alpha, \mu, \sigma, X_0) & \longmapsto J \big(\mathcal{L}(X^\alpha_{[0, T]}), \alpha\big).
    \end{align*}
\end{remark}

\begin{remark}
Under such a relaxation, the cost function $J$ in \eqref{eq: general_cost}  goes beyond the integral of a bounded continuous function, as usually assumed for standard  cost functions in the classical stochastic control theory. For example, consider the functional $J_0$ defined by:
\begin{equation*}
    \forall \big(\mathcal{L}(X_{[0, T]}), \alpha\big) \in \mathcal{P}\big(C([0, T], \R^n)\big) \times \A, \text{ } J_0 \big(\mathcal{L}(X_{[0, T]}), \alpha\big) := \E \left[ \int_0^T f(X_t, \alpha_t) \dd t + g(X_T) \right],
\end{equation*}
where $\alpha_t := \alpha(X_t)$, and $f: \R^n \times \R^k \to \R$ and $g: \R^n \to \R$ are given continuous functions that are of polynomial growth, \ie there exist $C > 0$ and $N \geq 0$ such that
\begin{equation*}
    \forall (x, y) \in \R^n \times \R^k, \quad |f(x, y)| < C(1 + ||x||^N + ||y||^N), \quad |g(x)| < C(1 + ||x||^N).
\end{equation*}

To see that $J_0$ satisfies the relaxed condition, let $\big\{(\alpha^{(i)}, \mu^{(i)}, \sigma^{(i)}, X_0^{(i)})\big\}_{i=0}^\infty$ be a sequence in the corresponding product space that converges to $(\alpha^{(\infty)}, \mu^{(\infty)}, \sigma^{(\infty)}, X_0^{(\infty)})$. Since $\{||X^{(i)}_t||^N\}_{i=0}^\infty$ is uniformly integrable for any $t \in [0, T]$, $\big\{|f(X^{(i)}_t, \alpha^{(i)}_t)|\big\}_{i=0}^\infty$ is also uniformly integrable. As a result, $\E\big(f(X^{(i)}_t, \alpha^{(i)}_t)\big)$ converges to $\E\big(f(X^{(\infty)}_t, \alpha^{(\infty)}_t)\big)$ as $i$ tends to infinity, \ie weak convergence + uniform integrability $\Longrightarrow$ convergence in expectation. The same argument applies to $g$. Finally, by dominated convergence theorem, the cost function value $J_0^{(i)}$ converges to $J_0^{(\infty)}$ as $i$ tends to infinity. Hence the continuity of $J_0$.

Moreover, the relaxed assumption include the following cost functions on the path space, which are beyond the classical setting where $f$ is a function of $(X_t, \alpha_t)$ and $g$ is a function of $X_T$:
\begin{itemize}
    \item Exotic option payoff:
    \[
    J \big(\mathcal{L}(X_{[0, T]}), \alpha\big) := \E \left[ \frac{1}{1 + \exp(-X_{T/2} \cdot X_T)} \right].
    \]
    
    \item Asian call option payoff:
    \[
    J \big(\mathcal{L}(X_{[0, T]}), \alpha\big) := \E\left[\left(\frac{1}{T} \int_0^T X_t \dd t - K\right)^+\right].
    \]
    
    \item Lookback call option payoff:
    \[
    J \big(\mathcal{L}(X_{[0, T]}), \alpha\big) := \E\left[\left(\max_{t \in [0, T]} X_t - K\right)^+\right].
    \]
\end{itemize}
\end{remark}

\section{Benefit of transfer learning: IPO and its local super-linear convergence for LQRs} \label{sec: LQR-transfer}

Having studied the general properties of transfer learning between continuous-time RL problems, we now present an explicit learning algorithm called IPO (\ie Iterative Policy Optimization) for the optimal policy of LQR, taking advantage of the Gaussian structure of its optimal policy.  

For this  IPO learning algorithm, we will first establish its global linear convergence, and then show its super-linear convergence when the initial policy lies in a certain neighborhood of the optimal policy. As a corollary, in the context of transfer learning,  such an algorithm yields an optimal policy for any closely related LQR with an appropriate initialization (\ie policy transfer). Our algorithm is analogous to the IPO algorithm developed for discrete-time LQRs in \cite{guo2026fast}, hence the adopted name IPO. 

The algorithm and the analysis rely crucially on the  Gaussian form of the LQR's optimal policy.  Indeed, given the special form of \eqref{eq: NE_policy}, it suffices to  only optimize within the following class of Gaussian policies:
\begin{equation} \label{eq: Gaussian_policy}
    \pi_t(\cdot \,|\, x) = \mathcal{N} (-K_t x,\, \Sigma_t),
\end{equation}
where $K_t$ and $\Sigma_t$ are of appropriate shapes, and there exists $\delta_1 > 0$ such that $\Sigma_t - \delta_1 I \succeq 0$ for any $t \in [0, T]$. By \eqref{eq: NE_policy}, we observe that
\begin{equation} \label{eq: NE_Gaussian}
    K_t^\ast = R^{-1}_t B_t^\dagger P_t, \quad \Sigma_t^\ast = \frac{\tau}{2} R_t^{-1}
\end{equation}
under the optimal policy of the LQR \eqref{eq: state} -- \eqref{eq: cost}.
First, we denote by $J^{K, \Sigma}$ the cost function associated with \eqref{eq: Gaussian_policy}, with
\begin{equation*}
    J^{K, \Sigma}(t, x) := \E_{u_s \sim \pi_s(\cdot \,|\, x_s)}\Bigg[ \int_t^T x_s^\dagger Q_s x_s + u_s^\dagger R_s u_s + \tau \log h_s (u_s \,|\, x_s) \dd s + x_T^\dagger Q^\prime x_T \,\Bigg|\, x_t = x \Bigg].
\end{equation*}
Next, by DPP, for any $\Delta t > 0$, $J^{K, \Sigma}(t, x)$ satisfies the Bellman equation:
\begin{align} \label{eq: prelim_IPO}
    J^{K, \Sigma}(t, x) = \E_{u \sim \pi_{K, \Sigma}} \bigg[ \int_t^{t + \Delta t} x_s^\dagger Q_s x_s + u_s^\dagger R_s u_s & + \tau \log h_s(u_s \,|\, x_s) \dd s \nonumber \\
    & + J^{K, \Sigma} (t + \Delta t, x_{t + \Delta t}) \, \Big| \, x_t = x \bigg],
\end{align}
that is, 
\newpage
\begin{align} \label{eq: Bellman}
    \pd{J^{K, \Sigma}}{t} + \big[(A_t - B_t K_t)x \big] \cdot \nabla J^{K, \Sigma} & + \frac{1}{2} (\sigma_t \sigma^\dagger_t) \cdot \Delta J^{K, \Sigma} \nonumber \\
    & + x^\dagger (Q_t + K_t^\dagger R_t K_t) x + \tr(\Sigma_t R_t) - \frac{\tau}{2} \log\big( (2\pi e)^k |\Sigma_t| \big) = 0,
\end{align}
with the terminal condition $J^{K, \Sigma} (T, x) = x^\dagger Q^\prime x$. By plugging in the ansatz 
\[
J^{K, \Sigma} (t, x) = x^\dagger P^K_t x + r^{K, \Sigma}_t,
\]
we obtain the coupled Riccati equations:
\begin{align}
    \frac{\dd P^{K}_t}{\dd t} + (A_t - B_t K_t)^\dagger P^{K}_t + P^{K}_t (A_t - B_t K_t) + Q_t + K_t^\dagger R_t K_t & = 0, \quad P^{K}_T = Q^\prime, \label{eq: P_t} \\
    \frac{\dd r^{K, \Sigma}_t}{\dd t} + \tr(\sigma_t^\dagger P^{K}_t \sigma_t + \Sigma_t R_t) - \frac{\tau}{2} \log\big( (2\pi e)^k |\Sigma_t| \big) & = 0, \quad r^{K, \Sigma}_T = 0. \label{eq: r_t_terminal}
\end{align}
Note that $P^K_t$ only depends on $K_t$, and $r^{K, \Sigma}_t$ depends on $(K_t, \Sigma_t)$. Recall that Assumption~\ref{ass: regular} is sufficient for \eqref{eq: P_t} to admit a unique $C^1$ solution taking values in $S^n_{\geq 0}$ (\cf \cite[Corollary~2.10]{yong2012stochastic}).

Now we can derive an IPO algorithm for updating the parameters in the Gaussian policy \eqref{eq: Gaussian_policy}, namely $K_t$ and $\Sigma_t$, with the goal of learning the parameters of the optimal randomized policy, which are denoted by $K^\ast_t$ and $\Sigma^\ast_t$ (\cf \eqref{eq: NE_Gaussian}).

\paragraph{Iterative policy optimization (IPO) derivation.}

Based on the Bellman equation~\eqref{eq: prelim_IPO}, we define the \textit{preliminary IPO algorithm} of $(K_t, \Sigma_t)$ by:
\begin{align*}
    K^{\rm prelim}_t, \Sigma^{\rm prelim}_t := \argmin_{\widetilde{K}, \widetilde{\Sigma}} \E_{u \sim \pi_{\widetilde{K}, \widetilde{\Sigma}}}\bigg[\int_t^{t + \Delta t} x_s^\dagger Q_s x_s + u_s^\dagger R_s u_s & + \tau \log h_s(u_s | x_s) \dd s \\
    & + J^{K, \Sigma} (t + \Delta t, x_{t + \Delta t}) \, \Big| \, x_t = x \bigg],
\end{align*}
which depends on the value of $\Delta t$ and is equivalent to:
\begin{align} \label{eq: IPO_def}
    K^{\rm prelim}_t, \Sigma^{\rm prelim}_t := \argmin_{\widetilde{K}, \widetilde{\Sigma}} \E_{u \sim \pi_{\widetilde{K}, \widetilde{\Sigma}}}\bigg[\frac{1}{\Delta t} \int_t^{t + \Delta t} & x_s^\dagger Q_s x_s + u_s^\dagger R_s u_s + \tau \log h_s (u_s \,|\, x_s) \dd s \nonumber \\
    & + \frac{1}{\Delta t} \Big[J^{K, \Sigma}(t + \Delta t, x_{t + \Delta t}) - J^{K, \Sigma}(t, x) \Big] \, \bigg| \, x_t = x \bigg].
\end{align}
Our \textit{IPO algorithm} is then defined by the limit of the above preliminary algorithm, that is, on the RHS of \eqref{eq: IPO_def}, we set $\Delta t \rightarrow 0^+$ and exchange the limit with $\argmin$ to obtain (\ie minimizing the first-order derivative of the RHS of \eqref{eq: prelim_IPO} at $\Delta t = 0$):
\begin{equation*}
    K^{\rm IPO}_t, \Sigma^{\rm IPO}_t := \argmin_{\widetilde{K}_t, \widetilde{\Sigma}_t} \bigg\{x^\dagger (\widetilde{K}_t^\dagger R_t \widetilde{K}_t - 2 \widetilde{K}_t^\dagger B_t^\dagger P^{K}_t) x + \tr(\widetilde{\Sigma}_t R_t) - \frac{\tau}{2} \log|\widetilde{\Sigma}_t| \bigg\},
\end{equation*}
which admits the following analytical solution:
\begin{align}
     K^{\rm IPO}_t & = R_t^{-1} B_t^\dagger P^{K}_t, \tag{IPO: $K$} \label{eq: IPO_K} \\
     \Sigma_t^{\rm IPO} & = \frac{\tau}{2} R^{-1}_t. \tag{IPO: $\Sigma$} \label{eq: IPO_Sigma}
\end{align}
where $P^{K}_t$ is the solution to \eqref{eq: P_t}. Notice that $\Sigma^{\rm IPO}_t$  reaches the covariance of the optimal Gaussian policy after a single iteration (\cf \eqref{eq: NE_Gaussian}).  We present below  the IPO algorithm for updating $K_t$.

\begin{algorithm}[H]
\begin{algorithmic}[1]
    \Require Initial value $(K^{(0)}_t)_{t \in [0, T]}$
    \State $i \gets 0$
    \While{not stop\_flag}
        \State Solve \eqref{eq: P_t} to obtain $(P^{K^{(i)}}_t)_{t \in [0, T]}$
        \State $K^{(i+1)}_t \gets R_t^{-1} B_t^\dagger P^{K^{(i)}}_t,\, t \in [0, T]$
        \State $i \gets i + 1$
    \EndWhile
    \State\Return $(K^{(i)}_t)_{t \in [0, T]}$
\end{algorithmic}
\caption{IPO algorithm for learning $(K^\ast_t)_{t \in [0, T]}$}
\end{algorithm}

\paragraph{Convergence of IPO.}

Now we show the convergence  of the IPO algorithm defined by \eqref{eq: IPO_K} -- \eqref{eq: IPO_Sigma}.  
We will show that with an additional  assumption  stated in Assumption \ref{ass: S_0} , the IPO algorithm admits global linear convergence. Since $(\Sigma^{\rm IPO}_t)_{t \in [0, T]}$ always reaches the covariance of the optimal Gaussian policy after a single iteration, we only discuss the convergence of $(K^{\rm IPO}_t)_{t \in [0, T]}$.

For any given parameters $(K_t, \Sigma_t)_{t \in [0, T]}$, we use the cost function value to measure their goodness (with an abuse of notation):
\begin{align} \label{eq: cost_function}
    C(K, \Sigma) & := J_{\pi_{K, \Sigma}} (0, \D_0) \\
    & = \E\left(x^\dagger P^K_0 x + r^{K, \Sigma}_0 \,\Big|\, x \sim \D_0 \right), \nonumber
\end{align}
where $(P^K_t, r^{K, \Sigma}_t)_{t \in [0, T]}$ solves the coupled Riccati equations \eqref{eq: P_t} -- \eqref{eq: r_t_terminal}. Note that $C(K, \Sigma)$ is minimized at $(K^\ast_t, \Sigma^\ast_t)_{t \in [0, T]}$ (\resp at $(K^\ast_t)_{t \in [0, T]}$ when viewed only as a functional in $K$). See \eqref{eq: NE_Gaussian} for the values of $(K^\ast_t, \Sigma^\ast_t)_{t \in [0, T]}$. 

\begin{assumption} \label{ass: S_0}
    $\E\big(x_0 x_0^\dagger \,\big|\, x_0 \sim \mathcal{D}_0 \big) \succ 0$.
\end{assumption}

\begin{proposition}[Global linear convergence of IPO] \label{prop: IPO_global}
    Under Assumptions~\ref{ass: prob} -- \ref{ass: S_0}, suppose that $\{ (K^{(i)}_t, \Sigma_t )_{t \in [0, T]} \}_{i \geq 0}$ is a sequence of parameters following the algorithm \eqref{eq: IPO_K}. Then, there exist constants $\mathcal{C}^K_1 > 0$ and $0 \leq \mathcal{C}_1 < 1$, which depend on $K^{(0)}$ and the data of the LQR \eqref{eq: state} -- \eqref{eq: cost}, such that $\forall i \geq 0$,
    \begin{equation*}
        \mathcal{C}^K_1 \int_0^T \left|\left|K^{(i + 1)}_t - K^\ast_t \right|\right|_2^2 \dd t \leq C (K^{(i+1)}, \Sigma) - C(K^\ast, \Sigma) \leq \mathcal{C}_1 \left[C (K^{(i)}, \Sigma) - C(K^\ast, \Sigma)\right].
    \end{equation*}
\end{proposition}

One can further establish the super-linear convergence for the IPO algorithm, with an appropriate initialization. 

\begin{proposition}[Local super-linear convergence of IPO] \label{prop: IPO_local}
    Under Assumptions~\ref{ass: prob} -- \ref{ass: S_0}, there exist constants $(\epsilon, \mathcal{C}_2) > 0$, which depend on the data of the LQR \eqref{eq: state} -- \eqref{eq: cost}, such that for any sequence of parameters $\{ (K^{(i)}_t, \Sigma_t )_{t \in [0, T]} \}_{i \geq 0}$ following the algorithm \eqref{eq: IPO_K} and satisfying:
    \[
    \int_0^T \left|\left|K^{(0)}_t - K^\ast_t\right|\right|_2^2 \dd t \leq \epsilon,
    \]
    the following local super-linear convergence holds:
    \begin{equation*}
        \forall i \geq 0, \quad C (K^{(i + 1)}, \Sigma) - C(K^\ast, \Sigma) \leq \mathcal{C}_2 \left[C (K^{(i)}, \Sigma) - C(K^\ast, \Sigma)\right]^{\frac{3}{2}}.
    \end{equation*}
\end{proposition}

\begin{remark}
    Assumption~\ref{ass: S_0} is critical in proving that the minimum eigenvalue of $\E(x_t x_t^\dagger)$ is uniformly bounded away from 0 (\cf Lemma~\ref{lemma: lower bound}). This uniform lower bound then leads to the uniform contraction of the IPO algorithm. In the discrete-time setting (\cf \cite{guo2026fast}), the counterpart of Assumption~\ref{ass: S_0} is also imposed to guarantee the global linear convergence of the algorithms (\cf \cite[Lemma~5.2]{guo2026fast}).
\end{remark}

\begin{remark}
    In fact, one can replace $\D_0$ with any square-integrable distribution in the definition of $C(\cdot, \cdot)$ (\cf \eqref{eq: cost_function}) and all the above convergence results still hold. This is because the initial distribution of the LQR \eqref{eq: state} -- \eqref{eq: cost} is irrelevant to the definition of the IPO algorithm. In this case, one only needs to change the statement of Assumption~\ref{ass: S_0} to guarantee the corresponding positive-definiteness.
\end{remark}

\paragraph{Transfer learning with IPO.}

Now combining Theorem~\ref{thm: general_TL} and Proposition~\ref{prop: IPO_local}, we immediately have the super-fast learning via appropriate policy transfer between LQRs. We mention that in Proposition~\ref{prop: IPO_local}, $\epsilon$ admits a lower bound which only depends on the norms of the LQR's model parameters.

\begin{corollary}[Transfer learning of LQRs with IPO] \label{thm: TL}
    Under Assumptions~\ref{ass: prob} -- \ref{ass: S_0}, denote by $(K^\ast_t)_{t \in [0, T]}$ the parameter of the optimal Gaussian policy of the LQR represented by $(A_{t \in [0, T]}, Q_{t \in [0, T]}, B_{t \in [0, T]}, R_{t \in [0, T]}, Q^\prime)$. Then, there exists $\epsilon > 0$, such that any initialization $(K^{(0)}_t)_{t \in [0, T]}$ converges super-linearly to the optimal Gaussian policy of any LQR represented by $(\tilde{A}_{t \in [0, T]}, \tilde{Q}_{t \in [0, T]}, \tilde{B}_{t \in [0, T]}, \tilde{R}_{t \in [0, T]}, \tilde{Q^\prime})$, provided that
    \begin{align*}
        ||K^{(0)} - K^\ast||_{2; [0, T]} + ||\tilde{A} - A||_{\infty;[0, T]} & + ||\tilde{Q} - Q||_{\infty;[0, T]} \\
        & + ||\tilde{B} - B||_{\infty;[0, T]} + ||\tilde{R} - R||_{\infty;[0, T]} + ||\tilde{Q^\prime} - Q^\prime||_2 < \epsilon,
    \end{align*}
    where $||\cdot||_{\infty;[0, T]}$ (\resp $||\cdot||_{2; [0, T]}$, $||\cdot||_2$) denotes the functional $L^\infty$ norm (\resp functional $L^2$ norm, matrix 2-norm).
\end{corollary}

\section{Application: stability of score-based diffusion models} \label{sec: application}

In this section, we will show that our analysis of LQRs, especially the Lemma~\ref{lemma: R_continuity} can be applied to analyze the stability of  score-based diffusion models. The critical observation is that the probability density function of (a certain class of) score-based diffusion models can be found in the LQRs under the optimal randomized policy. This allows us to consider a class of score matching functions and to bound the distance between the generated distribution and the target distribution.

\paragraph{Mechanism of score-based diffusion models.}

Score-based diffusion models have become the SOTA solution to various tasks in different areas. For completeness, we first recall their basic mechanism briefly. (See for example \cite{tang2025score} for a  comprehensive review). 

\begin{figure}[H] 
\centering
\begin{tikzpicture}[->, >=latex]
    \node (a) at (0,0) {(4) $p^{\rm data}_\approx$};
    \node (b) at (6,0) {(3) $p^{\rm noise}$};
    \node (c) at (0,2) {(1) $p^{\rm data}_0$};
    \node (d) at (6,2) {(2) $p^{\rm data}_T$};
    
    \path[->] (b) edge node[below] {$e_2 := d(s_\beta, s)$} (a);
    \path[->] (b) edge node[above] {\small Backward SDE} (a);
    \path[->] (c) edge node[above] {\small Forward SDE} (d);
    \path[->] (d) edge node[right] {$e_1 := d\left(p^{\rm noise}, p^{\rm data}_T\right)$} (b);
    \path[->] (a) edge node[left] {$e_3 \in O(e_1 + e_2)$} (c);
\end{tikzpicture}
\caption{Basic mechanism of score-based diffusion models.}
\label{fig: diffusion}
\end{figure}
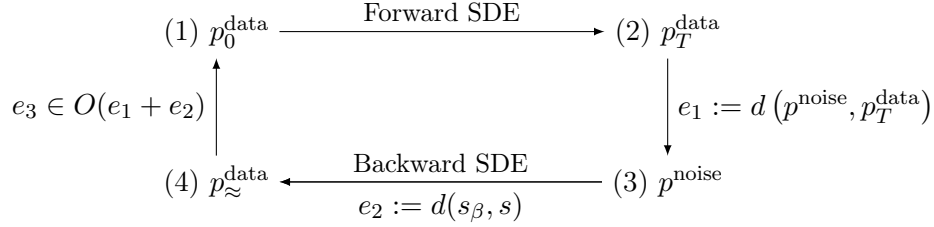

Suppose $p^{\rm data}_0$ is the distribution that one aims to generate. Diffusion model starts by defining a forward SDE such as an O-U process over $[0, T]$ with the initial distribution $p^{\rm data}_0$. Denote by $s$ and $p^{\rm data}_T$ the score function and the terminal distribution of the forward SDE, respectively. Then, in the sampling stage, a backward SDE, whose dynamics depend on $s$ and whose initial distribution is $p^{\rm data}_T$, is simulated. 
(Figure~\ref{fig: diffusion} summarizes the basic mechanism of score-based diffusion models). Theoretically, it can be shown that the terminal distribution of the backward SDE is equal to $p^{\rm data}_0$. In practice, however, $s$ and $p^{\rm data}_T$ are typically not accessible, and a score matching function $s_\beta$ and a noise distribution $p^{\rm noise}$ are adopted as their approximations, respectively. Denote by $p^{\rm data}_\approx$ the terminal distribution of the backward SDE (under $s_\beta$ and $p^{\rm noise}$). 

\paragraph{Connection with LQRs.}

Next, we recall the connection between score-based diffusion models and LQRs, stated in Lemma~\ref{lemma: diffusion}, which  can be viewed as a special case of \cite[Theorem~7]{zhang2023mean}. The key ingredient is the Cole-Hopf transformation for the HJB equation that characterizes the optimal policy of the LQR (\cf \eqref{eq: HJB} in our case).

\begin{lemma} \label{lemma: diffusion}
Take the LQR \eqref{eq: state} -- \eqref{eq: cost} with Assumptions~\ref{ass: prob} -- \ref{ass: regular}. Further assume
    \begin{equation*}
        \tr(A_t)= - \frac{\tau}{4} \log\frac{|R_t|}{(\tau \pi)^k}, \quad B_t R_t^{-1} B_t^\dagger = \sigma_t \sigma_t^\dagger, \quad Q_t = 0
    \end{equation*}
    for any $t \in [0, T]$, and $Q^\prime \succ 0$.
    Then the probability density function $\hat{p}(t, x)$ of the following diffusion process on $[0, T]$:
    \begin{equation} \label{eq: diffusion_SDE}
        \dd \hat{X}_t = -A_{T - t} \hat{X}_t \dd t + \sigma_{T - t} \dd W_t, \quad \hat{X}_0 \sim \mathcal{N} \left(0, (Q^\prime)^{-1} \right)
    \end{equation}
    can be expressed as
    \begin{equation*}
        \hat{p}(t, x) = (2\pi)^{-\frac{n}{2}} |Q^\prime|^\frac{1}{2} \exp \left[ -\frac{1}{2} J(T - t, x) \right],
    \end{equation*}
    where $J(t, x) = x^\dagger P_t x + r_t$ with $(P_t, r_t)$ solving the coupled Riccati equations \eqref{eq: P_t_1} -- \eqref{eq: r_t_2}. 
\end{lemma}


We note that \eqref{eq: diffusion_SDE} specifies a diffusion model where the data distribution $\hat{X}_0$ (\ie the distribution one aims to generate) is Gaussian, and the forward SDE is an O-U process. By Lemma~\ref{lemma: diffusion}, $\hat{p}(t, x)$ is determined by $P_{T - t}$ and $r_{T - t}$. In fact, $\hat{p}(t, x)$ is determined solely by $P_{T - t}$ since the spacial integral of $\hat{p}(t, x)$ must be 1. As a result, the score function of \eqref{eq: diffusion_SDE} (\ie the gradient of $\log \hat{p}(t, x)$) is determined by $P_{T - t}$.

\paragraph{A class of score matching functions and the stability.}

Now, the backward SDE of \eqref{eq: diffusion_SDE} is:
\begin{equation} \label{eq: diffusion_backward}
    \dd \hat{Y}_t = \left[A_t \hat{Y}_t + \sigma_t \sigma_t^\dagger \nabla \log \hat{p}^{Q^\prime}(T - t, \hat{Y}_t)\right] \dd t + \sigma_t \dd W_t, \quad \hat{Y}_0 \sim \hat{p}^{Q^\prime}(T, \cdot),
\end{equation}
where we use $\hat{p}^{Q^\prime}$ to indicate the dependence of $\hat{p}$ on $Q^\prime$. In practice, when $\hat{p}^{Q^\prime}$ is not explicitly known, a score matching function $s_\beta$ is used as an approximator of the score function $\nabla \log \hat{p}^{Q^\prime}$, and the initial distribution is approximated by some noise distribution $p^{\rm noise}$, \ie
\begin{equation} \label{eq: diffusion_approx}
    \dd Y_t = \left[A_t Y_t + \sigma_t \sigma_t^\dagger s_\beta(T - t, Y_t) \right] \dd t + \sigma_t \dd W_t, \quad Y_0 \sim p^\text{noise}.
\end{equation}
We will show that $Y_T \approx \hat{Y}_T \overset{d}{=} \hat{X}_0$ when $s_\beta \approx \nabla \log \hat{p}^{Q^\prime}$ and $p^\text{noise} \approx \hat{p}^{Q^\prime}(T, \cdot)$: this follows from  the stability of the Riccati equation~\eqref{eq: P_t_1} (\cf Lemma~\ref{lemma: R_continuity}), such that  $s_\beta = \nabla \log \hat{p}^M$ serves as a good score matching function as long as $M \approx Q^\prime$.
 
\begin{theorem}[Error bound analysis]
    Take the same setting of Lemma \ref{lemma: diffusion}. Then, there exist constants $(C_1, C_2, C_3) > 0$, which depend on the data of the LQR \eqref{eq: state} -- \eqref{eq: cost}, such that for any $\epsilon > 0$, there exists $\delta_0 > 0$, such that $||M - Q^\prime|| < \delta_0$ implies 
    \[
    d_{\rm TV} \left(Y_T, \hat{Y}_T\right) \leq d_{\rm TV} \left(p^{\rm noise}, \hat{p}^{Q^\prime} (T, \cdot)\right) + C_1 \epsilon,
    \]
    and
    \[
    W_2 \left(Y_T, \hat{Y}_T\right) \leq \sqrt{C_2 W_2^2 \left(p^{\rm noise},\hat{p}^{Q^\prime} (T, \cdot)\right) + C_3 \epsilon^2},
    \]
    where $Y_t$ satisfies \eqref{eq: diffusion_approx} with $s_\beta = \nabla \log \hat{p}^M$, and $\hat{Y}_t$ satisfies \eqref{eq: diffusion_backward}. Here $d_{\rm TV}$ and $W_2$ to denote the total variation distance and 2-Wasserstein distance, respectively.
\end{theorem}

\begin{proof}
    Our proof utilizes the results in \cite[Section~6]{tang2025score}. We first prove the total variation bound. By Lemma~\ref{lemma: R_continuity}, for any fixed $x \in \R^n$, we have
    \[
    \nabla \log q^M (\cdot, x) \rightarrow \nabla \log q^{Q^\prime} (\cdot, x) \text{ in } C([0, T], \R^n)
    \]
    as $M \rightarrow Q^\prime$. Then, we have 
    \[
    \forall t \in [0, T], \quad \E_{\hat{X}_t \sim q(t, \cdot)} \Big|\Big| \nabla \log q^M (t, \hat{X}_t) - \nabla \log q^{Q^\prime} (t, \hat{X}_t) \Big|\Big|^2 \rightarrow 0
    \]
    as $M \rightarrow Q^\prime$. The total variation bound is then proved by invoking \cite[Theorem~6.2]{tang2025score}. Similarly, the 2-Wasserstein bound can be proved by invoking \cite[Theorem~6.5]{tang2025score}.
\end{proof}

\section{Proof of lemmas and propositions} \label{sec: proofs}

\subsection{Proof of Lemma \ref{lemma:optimalpolicy}}

\begin{proof}
    Define the following intermediate cost function:
    \begin{equation} \label{eq: intermediate_cost}
        J(t, x) := \inf_{\pi \in \A} \E_{u_s \sim \pi_s(\cdot \,|\, x_s)}\Bigg[ \int_t^T x_s^\dagger Q_s x_s + u_s^\dagger R_s u_s + \tau \log h_s (u_s \,|\, x_s) \dd s + x_T^\dagger Q^\prime x_T \,\Bigg|\, x_t = x \Bigg].
    \end{equation}
    Then, DPP produces the following HJB equation of the LQR \eqref{eq: state} -- \eqref{eq: cost}:
    \begin{align} 
        -\pd{J(t, x)}{t} = \inf_{\pi \in \A} \E_{u_t \sim \pi_t} \Bigg\{ & (A_t x + B_t u_t) \cdot \nabla J(t, x) + \frac{1}{2} (\sigma_t \sigma_t^\dagger) \cdot \Delta J(t, x) \nonumber \\ \tag{HJB} \label{eq: HJB}
        & + x^\dagger Q_t x + u_t^\dagger R_t u_t + \tau \log h_t (u_t \,|\, x) \Bigg\}, \quad J(T, x) = x^\dagger Q^\prime x, 
    \end{align}
    where $h_t (\cdot \,|\, x)$ denotes the (conditional) probability distribution function of the Markov randomized policy $\pi_t(\cdot \,|\, x)$.
    
    By \cite[Lemma 2.2]{guo2026fast}, the RHS of \eqref{eq: HJB} is minimized by the following Gaussian policy:
    \begin{equation*}
        \pi^\ast_t (\cdot \,|\, x) = \mathcal{N} \left(-\frac{1}{2} R_t^{-1} B_t^\dagger \nabla J(t, x),\, \frac{\tau}{2} R_t^{-1} \right).
    \end{equation*}
    Observing the linear quadratic nature of LQRs, we introduce the ansatz for $J (t, x)$ such that 
    \[
    J(t, x) = x^\dagger P_t x + r_t.
    \]
    After plugging the ansatz for $J$ and the expression of $\pi^\ast$ into \eqref{eq: HJB}, we obtain the following coupled Riccati equations for $(P_t, r_t)$:
    \begin{align}
        \frac{\dd P_t}{\dd t} + A_t^\dagger P_t + P_t A_t + Q_t - P_t B_t R^{-1}_t B_t^\dagger P_t & = 0, \quad P_T = Q^\prime, \tag{\ref{eq: P_t_1}} \\
        \frac{\dd r_t}{\dd t} + \tr(\sigma_t^\dagger P_t \sigma_t) + \frac{\tau}{2} \log\frac{|R_t|}{(\tau \pi)^k} & = 0, \quad r_T = 0. \label{eq: r_t_2}
    \end{align}
    Hence the lemma. 
\end{proof}

\subsection{Proof of Lemma \ref{lemma: SDE_stability}}

\begin{proof}
    In this proof, we use $\hat{\mathcal{R}}(\mu, \sigma, Y_0)$ to denote the RDE solution of \eqref{eq: RDE_Strat} as a path-valued random variable (\cf Lemma~\ref{lemma: RDE}). To clarify, it holds that $\mathcal{R} = \mathcal{L} \circ \hat{\mathcal{R}}$. 
    
    Suppose $(\mu^1, \sigma^1, y^1_0) \in {\rm Lip}^1 (\R^n, \R^n) \times {\rm Lip}^2 (\R^n, \R^{n \times d}) \times \R^n$. First, according to Lemma~\ref{lemma: RDE}, \cite[Theorem 12.11]{friz2010multidimensional} (\ie the stability of RDEs) and the fact that BM has a finite $p$-variation almost surely for any $p > 2$, it is not hard to see that for any $\epsilon > 0$ and $\eta \in [0, 1)$, there exists $\zeta(\epsilon, \eta) > 0$, such that for any $(\mu^2, \sigma^2, y^2_0) \in {\rm Lip}^1 (\R^n, \R^n) \times {\rm Lip}^2 (\R^n, \R^{n \times d}) \times \R^n$ with
    \begin{equation} \label{eq: measurability_proof}
        ||\mu^2 - \mu^1||_{\infty} + ||\sigma^2 - \sigma^1||_{\rm{Lip}^1} + ||y^2_0 - y^1_0|| < \zeta(\epsilon, \eta),
    \end{equation}
    we have
    \begin{equation} \label{eq: convergence_prob}
        \mathbb{P} \Big(||\hat{\mathcal{R}} (\mu^2, \sigma^2, y_0^2) - \hat{\mathcal{R}} (\mu^1, \sigma^1, y_0^1)||_{\infty; [0, T]} < \epsilon \Big) \geq \eta.
    \end{equation}
    
    Second, since $||\hat{R}(\mu^1, \sigma^1, y^1_0)||_{\infty;[0, T]}$ is almost surely finite and its tail distribution only depends on the value of $||\mu^1||_{\infty} + ||\sigma^1||_{\rm{Lip}^1} + ||y^1_0||$, the condition in \eqref{eq: measurability_proof} can be weakened to
    \[
    ||\mu^2 - \mu^1||_{\infty; \text{compact}} + ||\sigma^2 - \sigma^1||_{\rm{Lip}^1; \text{compact}} + ||y^2_0 - y^1_0|| < \zeta(\epsilon, \eta).
    \]
    
    Third, let $Y^1_0$ and $Y^2_0$ be two square-integrable $\R^n$-valued random variables. We see from the Markov inequality
    \[
    \mathbb{P}\big(||Y^1_0 - Y^2_0|| > \epsilon \big) \leq \frac{||Y^1_0 - Y^2_0||^2_{L^2}}{\epsilon^2}
    \]
    that large deviations of $Y_0^1 - Y^2_0$ can be bounded by its $L^2$ norm. Therefore, \eqref{eq: convergence_prob} holds when $y^1_0$ and $y^2_0$ are two square-integrable random variables. One only needs to replace $||y^1_0 - y^2_0||$ in \eqref{eq: measurability_proof} by $||y^1_0 - y^2_0||_{L^2}$. 

    At last, we notice from \eqref{eq: convergence_prob} that $\hat{\mathcal{R}} (\mu^2, \sigma^2, y_0^2)$ converges to $\hat{\mathcal{R}} (\mu^1, \sigma^1, y_0^1)$ in probability as random variables on the metric space $C([0, T], \R^n)$. Since convergence in probability implies weak convergence (see \eg \cite[Section 3]{billingsley2013convergence}), we conclude that $\hat{\mathcal{R}} (\mu^2, \sigma^2, y_0^2)$ converges weakly to $\hat{\mathcal{R}} (\mu^1, \sigma^1, y_0^1)$. This proves the continuity of $\mathcal{R}$.
\end{proof}

\subsection{Proof of Proposition~\ref{prop: IPO_global} (global linear convergence of IPO)}

In the following, with an abuse of notation, we sometimes use $\langle \cdot, \cdot \rangle$ to indicate the usual matrix inner product. In addition, for any matrix $M$, we use $\lambda_{\min} (M)$ (\resp $||M||_2$) to denote the square root of the smallest (\resp largest) eigenvalue of $M^\dagger M$. 

We first define the following matrix-valued functions. 
\begin{align*}
    \G(t, K^\prime, K) & := P^K_t \big[B_t (K_t - K^\prime_t)\big] + \big[B_t (K_t - K^\prime_t)\big]^\dagger P^K_t + {K^\prime}^\dagger_t R_t K^\prime_t - K^\dagger_t R_t K_t, \\
    G(t, K) & := - \G(t, R^{-1} B^\dagger P^K, K) = P^K_t B_t R^{-1}_t B^\dagger_t P^K_t + K^\dagger_t R_t K_t - P^K_t B_t K_t - K^\dagger_t B^\dagger_t P^K_t.
\end{align*}
Since $K_t^\ast = R^{-1}_t B_t^\dagger P_t^\ast$, we have $\G(t, K, K^\ast) = (K_t - K^\ast_t)^\dagger R_t (K_t - K^\ast_t) \succeq 0$. Also, it can be verified by algebraic calculation that $G(t, K) \succeq 0$. In addition, for notational simplicity, in the rest of this section, we define
\begin{align} \label{eq: IPO_y_t} \tag{\ding{65}}
    y_t := \E(x_t x_t^\dagger), 
\end{align}
where $x_t$ solves the state SDE \eqref{eq: state} with $u_t$ following the policy $\pi_t = \mathcal{N}(-K_t x, \Sigma_t)$. And we shall use superscripts to indicate different policies. For instance, by $y^\prime_t$ we imply that $y^\prime_t = \E\left(x^\prime_t (x^\prime_t)^\dagger\right)$ where $x^\prime_t$ solves the state SDE \eqref{eq: state} with $u_t$ following the policy $\pi_t = \mathcal{N}(-K^\prime_t x, \Sigma_t)$. 

The proof of the global linear convergence relies on the following lemmas.

\begin{lemma}[Cost difference] \label{lemma: cost difference}
    Under Assumptions~\ref{ass: prob} -- \ref{ass: regular}, the cost difference of two parametrized Gaussian policies is given by:
    \begin{equation*}
        C(K^\prime, \Sigma) - C(K, \Sigma) = \int_0^T \Big\langle y^\prime_t, \G(t, K^\prime, K) \Big\rangle \dd t,
    \end{equation*}
    where $y^\prime_t$ is defined by \eqref{eq: IPO_y_t}.
\end{lemma}

\begin{proof}
    For notational simplicity, denote $J^\prime (t, x) := J_{K^\prime, \Sigma}(t, x)$ and $J (t, x) := J^{K, \Sigma} (t, x)$. By subtracting the two Bellman equations that $J^\prime (t, x)$ and $J(t, x)$ satisfy (\cf \eqref{eq: Bellman}), we obtain 
    \begin{equation*}
        \pd{(J^\prime - J)}{t} + \big[(A_t - B_t K_t^\prime) x\big] \cdot \nabla (J^\prime - J) + \frac{1}{2} (\sigma_t \sigma_t^\dagger) \cdot \Delta (J^\prime - J) + F(t, x) = 0,
    \end{equation*}
    where 
    \[
    F(t, x) = \big[B_t(K_t - K_t^\prime) x\big] \cdot \nabla J + (K^\prime x)^\dagger R (K^\prime x) - (Kx)^\dagger R (Kx).
    \]
    Now, define $u(t, x) := J^\prime (t, x) - J(t, x)$, then  by It\^o's formula, $
        \E \big[\dd u(t, x_t^\prime)\big] = \E\big[F(t, x^\prime_t)\big] \dd t,$
    where $x_t^\prime$ solves the state SDE \eqref{eq: state} with $u_t$ following the policy $\pi^\prime_t = \mathcal{N}(-K^\prime_t x, \Sigma_t)$. Finally, by integrating on $[0, T]$, we have:
    \begin{equation*}
        C(K^\prime, \Sigma) - C(K, \Sigma) = -\E \left[\int_0^T \dd u(t, x_t^\prime)\right] = \E \left[\int_0^T F(t, x^\prime_t) \dd t \right].
    \end{equation*}
    A manipulation of the matrices finishes the proof.
\end{proof}

\begin{lemma}[Contraction of IPO] \label{lemma: contraction}
    Under Assumptions~\ref{ass: prob} -- \ref{ass: S_0}, suppose $K^\prime$ is the one-step update of $K$ following the algorithm \eqref{eq: IPO_K}. Then, 
    \begin{equation*}
        C(K^\prime, \Sigma) - C(K^\ast, \Sigma) \leq \left\{1 - \frac{\min_{t \in [0, T]}\lambda_{\min}(y^\prime_t)}{\max_{t \in [0, T]} ||y^\ast_t||_2}\right\} \big[C(K, \Sigma) - C(K^\ast, \Sigma)\big],
    \end{equation*}
    where $K^\ast$ is the parameter of the optimal policy, and $y_t^\prime$ (\resp $y^\ast_t$) is defined by \eqref{eq: IPO_y_t}.
\end{lemma}

\begin{proof}
    By Lemma~\ref{lemma: cost difference}, we have:
    \begin{align*}
        C(K, \Sigma) - C(K^\ast, \Sigma) & = -\int_0^T \Big\langle y^\ast_t, \G(t, K^\ast, K) \Big\rangle \dd t
         \leq \int_0^T \Big\langle y^\ast_t, G(t, K) \Big\rangle \dd t \\
        & \leq \int_0^T ||y^\ast_t||_2 \tr\big[G(t, K)\big] \dd t \\
        & \leq \max_{t \in [0, T]} ||y^\ast_t||_2 \int_0^T \tr\big[G(t, K)\big] \dd t.
    \end{align*}
    Moreover,
    \begin{align*}
        C(K^\prime, \Sigma) - C(K, \Sigma) & = -\int_0^T \Big\langle y^\prime_t, G(t, K) \Big\rangle \dd t \\
        & \leq -\min_{t \in [0, T]} \lambda_{\min}(y^\prime_t) \int_0^T \tr\big[G(t, K)\big] \dd t \\
        & \leq -\frac{\min_{t \in [0, T]}\lambda_{\min}(y^\prime_t)}{\max_{t \in [0, T]} ||y^\ast_t||_2} \big[C(K, \Sigma) - C(K^\ast, \Sigma)\big].
    \end{align*}
    Adding $C(K, \Sigma) - C(K^\ast, \Sigma)$ to both sides gives the desired result.
\end{proof}

\begin{lemma}[Lower bound of $\lambda_{\min}$] \label{lemma: lower bound}
    Under Assumptions~\ref{ass: prob} -- \ref{ass: S_0}, suppose $\left\{ \left(K^{(i)}, \Sigma\right) \right\}_{i \geq 0}$ is a sequence of parameters following the algorithm \eqref{eq: IPO_K}. Then, there exists $\underline{\mu} > 0$, which is affected by $K^{(0)}$, such that:
    \begin{equation*}
        \forall i \geq 0,\, t \in [0, T], \quad \lambda_{\min} (y^{(i)}_t) \geq \underline{\mu},
    \end{equation*}
    where $y_t^{(i)}$ is defined by \eqref{eq: IPO_y_t}.
\end{lemma}

\begin{proof}
    For any fixed Gaussian policy parameterized by $(K, \Sigma)$, $y_t$ follows the ODE from It\^o's formula:
    \begin{equation*}
        \frac{\dd y_t}{\dd t} = (A_t - B_t K_t) y_t + y_t (A_t - B_t K_t)^\dagger + \sigma_t \sigma^\dagger_t, \quad y_0 = \E (x_0 x_0^\dagger \,|\, x_0 \sim \mathcal{D}_0 ).
    \end{equation*}
    Noticing that $\sigma_t \sigma^\dagger_t \succeq 0$, by adapting the proof of \cite[Lemma 3.7]{giegrich2024convergence}, we obtain:
    \begin{equation}
        \min_{t \in [0, T]}\lambda_{\min}(y_t) \geq \lambda_{\min} (y_0) \exp\left(-2 \int_0^T ||A_t - B_t K_t||_2 \dd t\right). \label{eq: inf_sigma}
    \end{equation}
    
    For the sequence of parameters $\left\{ \left(K^{(i)}, \Sigma\right) \right\}_{i \geq 0}$ defined by \eqref{eq: IPO_K}, we define $\Delta P^{(i)} := P^{K^{(i + 1)}} - P^{K^{(i)}}$. It satisfies the Riccati equation (\cf \eqref{eq: P_t}):
    \begin{equation*}
        \frac{\dd \Delta P^{(i)}}{\dd t} + (A_t - B_t K^{(i + 1)}_t)^\dagger \Delta P^{(i)} + \Delta P^{(i)} (A_t - B_t K^{(i + 1)}_t) = G(t, K^{(i)}), \quad \Delta P^{(i)}_T = 0.
    \end{equation*}
    Since $G(t, K^{(i)}) \succeq 0$, it implies that $\Delta P^{(i)} \preceq 0$ (\cf the proof of \cite[Proposition 3.5(1)]{giegrich2024convergence}). Therefore, for any $i \geq 1$, $||K^{(i)}_t||_2 \leq ||R^{-1}_t B^\dagger_t||_2 ||P^{K^{(0)}}_t||_2$, \ie the matrix 2-norm is bounded from above. Combining this upper bound with \eqref{eq: inf_sigma} yields the desired conclusion.
\end{proof}

The following lemma is immediate from Lemma~\ref{lemma: cost difference} and Lemma~\ref{lemma: lower bound}, and by observing that:
\[
\G(t, K, K^\ast) = (K_t - K^\ast_t)^\dagger R_t (K_t - K^\ast_t) \succeq 0.
\]

\begin{lemma}[Upper bound of $L^2$ distance] \label{lemma: IPO_K_diff}
    Under Assumptions~\ref{ass: prob} -- \ref{ass: S_0}, suppose $\left\{ \left(K^{(i)}, \Sigma\right) \right\}_{i \geq 0}$ is a sequence of parameters following the algorithm \eqref{eq: IPO_K}. Then,
    \[
    \forall i \geq 0, \quad C(K^{(i)}, \Sigma) - C(K^\ast, \Sigma) \geq \underline{\mu} \delta \int_0^T ||K^{(i)}_t - K^\ast_t ||_2^2 \dd t,
    \]
    where $\underline{\mu} > 0$ is defined in Lemma~\ref{lemma: lower bound}, and $\delta > 0$ is defined in Assumption~\ref{ass: regular}.
\end{lemma}

\subsection{Proof of Proposition~\ref{prop: IPO_local} (local super-linear convergence of IPO)} \label{sec: IPO_proof}

The proof of local super-linear convergence is built upon a series of lemmas.

\begin{lemma}[Contraction of IPO] \label{lemma: super 1}
    Under Assumptions~\ref{ass: prob} -- \ref{ass: S_0}, suppose $\left\{ \left(K^{(i)}, \Sigma\right) \right\}_{i \geq 0}$ is a sequence of parameters following the algorithm \eqref{eq: IPO_K} and satisfying 
    \[
    \forall i \geq 1, \quad \max_{t \in  [0, T]} ||y^{(i)}_t - y^\ast_t ||_2 \leq \min_{t \in [0, T]} \lambda_{\min}(y^\ast_t).
    \]
    Then,
    \begin{equation*}
        C (K^{(i + 1)}, \Sigma) - C(K^\ast, \Sigma) \leq \frac{\max_{t \in  [0, T]} ||y^{(i + 1)}_t - y^\ast_t ||_2}{\min_{t \in [0, T]}\lambda_{\min}(y^\ast_t)} \left[C(K^{(i)}, \Sigma) - C(K^\ast, \Sigma)\right].
    \end{equation*}
\end{lemma}

\begin{proof}
    Denote $K^\prime := K^{(i + 1)}$ and $K := K^{(i)}$ to simplify the notation. By Lemma~\ref{lemma: cost difference}, we obtain:
    \begin{align*}
        C(K^\prime, \Sigma) - C(K, \Sigma) & = \int_0^T \Big\langle y^\prime_t, \G(t, K^\prime, K) \Big\rangle \dd t 
         = -\int_0^T \Big\langle y^\prime_t, G(t, K) \Big\rangle \dd t \\
        & = -\int_0^T \Big\langle y^\ast_t, G(t, K) \Big\rangle - \Big\langle y^\prime_t - y^\ast_t, G(t, K) \Big\rangle \dd t\\
        & \leq \left(-1 + \frac{\max_{t \in  [0, T]}||y^\prime_t - y^\ast_t||_2}{\min_{t \in [0, T]}\lambda_{\min}(y^\ast_t)}\right) \int_0^T \Big\langle y^\ast_t, G(t, K) \Big\rangle \dd t \\
        & \leq \left(-1 + \frac{\max_{t \in  [0, T]}||y^\prime_t - y^\ast_t||_2}{\min_{t \in [0, T]}\lambda_{\min}(y^\ast_t)}\right) \Big[C(K, \Sigma) - C(K^\ast, \Sigma)\Big].
    \end{align*}
    Adding $C(K, \Sigma) - C(K^\ast, \Sigma)$ to both sides gives the desired result.
\end{proof}

\begin{lemma}[Perturbation of $y_t$] \label{lemma: super 2}
    Let $\rho > 0$. Under Assumptions~\ref{ass: prob} -- \ref{ass: regular}, suppose the two policies $\left\{\left(K^i, \Sigma\right)\right\}_{i=1, 2}$ satisfy 
    \begin{equation*}
        \max_{1 \leq i \leq 2} \int_0^T \left|\left|A_t - B_t K^i_t\right|\right|_2 \dd t \leq \rho.
    \end{equation*}
    Then, there exists $\hat{c}_\rho > 0$ such that 
    \begin{equation*}
        \max_{t \in [0, T]} \left|\left|y^1_t - y^2_t\right|\right|_2 \leq \hat{c}_\rho \int_0^T \left|\left|K^1_t - K^2_t\right|\right|_2 \dd t.
    \end{equation*}
\end{lemma}

\begin{proof}
We divide the proof into several steps.

\noindent
\fbox{\textit{Step 1: Calculate the perturbation of $y_t$.}}

By It\^o's formula, $y_t$ satisfies the ODE:
\begin{equation}
    \frac{\dd y_t}{\dd t} = (A_t - B_t K_t) y_t + y_t (A_t - B_t K_t)^\dagger + \sigma_t \sigma_t^\dagger, \quad y_0 = \E (x_0 x_0^\dagger \,|\, x_0 \sim \mathcal{D}_0 ). \label{eq: y_t_ode}
\end{equation}
By subtracting the ODEs that $y^1_t$ and $y^2_t$ satisfy, we get:
\begin{align} \label{eq: y_t_diff}
    \frac{\dd (y^1_t - y^2_t)}{\dd t} = (A_t - B_t K^1_t) (y^1_t - y^2_t) & + (y^1_t - y^2_t) (A_t - B_t K^1_t)^\dagger \nonumber \\
    & - \big[B_t(K^1_t - K^2_t)\big] y^2_t - y^2_t \big[B_t (K^1_t - K^2_t)\big]^\dagger, 
\end{align}
with the initial condition $y^1_0 - y^2_0 = 0$.

\noindent
\fbox{\textit{Step 2: Bound the norm of $y_t$.}}

By integrating over $[0, t]$ and then taking norms on both sides of \eqref{eq: y_t_ode}, we get:
\begin{equation*}
    ||y_t||_2 \leq ||y_0||_2 + 2 \int_0^t ||A_s - B_s K_s||_2 ||y_s||_2 + ||\sigma_s \sigma_s^\dagger||_2 \dd s.
\end{equation*}
By Gronwall's inequality, there exists $\tilde{c}_\rho > 0$ such that $
    \max_{t \in [0, T]} ||y_t||_2 \leq \tilde{c}_\rho.$
\noindent
\fbox{\textit{Step 3: Bound the perturbation of $y_t$.}}

By integrating over $[0, t]$ and then taking norms on both sides of \eqref{eq: y_t_diff}, we get
\begin{align*}
    ||y^1_t - y^2_t||_2 & \leq 2 \int_0^t ||A_s - B_s K^1_s||_2 ||y^1_s - y^2_s||_2 + ||B_s||_2 ||K^1_s - K^2_s||_2 ||y^2_s||_2 \dd s \\
    & \leq 2 \int_0^t ||A_s - B_s K^1_s||_2 ||y^1_s - y^2_s||_2 \dd s + 2 \tilde{c}_\rho \max_{t \in [0, T]}||B_t||_2 \int_0^t ||K^1_s - K^2_s||_2 \dd s.
\end{align*}
Then by Gronwall's inequality, there exists $\hat{c}_\rho > 0$ such that
\[
\max_{t \in [0, T]} ||y^1_t - y^2_t||_2 \leq \hat{c}_\rho \int_0^T ||K^1_t - K^2_t||_2 \dd t.
\]
\end{proof}

\begin{lemma}[Bound the one-step update of $y_t$] \label{lemma: super 3}
    Under Assumptions~\ref{ass: prob} -- \ref{ass: regular}, let $\rho > 0$ be such that 
    \[
    \int_0^T ||A_t - B_t K^\ast_t||_2 \dd t \leq \rho.
    \]
    Suppose $\left\{ \left(K^{(i)}, \Sigma\right) \right\}_{i \geq 0}$ is a sequence of parameters following the algorithm \eqref{eq: IPO_K} and satisfying
    \[
    \sup_{i \geq 0} \int_0^T ||A_t - B_t K^{(i)}_t||_2 \dd t \leq \rho.
    \] 
    Then, there exists $c_\rho^\ast > 0$ which is affected by $K^{(0)}$, such that for any $i \geq 0$, we have
    \begin{equation*}
        \forall i \geq 0, \quad \max_{t \in [0, T]}||y^{(i + 1)}_t - y^\ast_t||_2 \leq c_\rho^\ast \int_0^T ||K^{(i)}_t - K^\ast_t||_2 \dd t.
    \end{equation*}
\end{lemma}

\begin{proof}
    Denote $K^\prime := K^{(i + 1)}$ and $K := K^{(i)}$. Then, by definition,
    \begin{equation*}
        ||K^\prime_t - K^\ast_t||_2 = ||R^{-1}_t B^\dagger_t (P^K_t - P^{K^\ast}_t)||_2  \leq \frac{||B_t||_2}{\lambda_{\min}(R_t)} ||P^K_t - P^{K^\ast}_t||_2.
    \end{equation*}
    Note that $P^K_t - P^{K^\ast}_t$ satisfies the ODE:
    \begin{align*}
        \frac{\dd (P^K_t - P^{K^\ast}_t)}{\dd t} = (A_t - B_t K_t)^\dagger & (P^K_t - P^{K^\ast}_t) + (P^K_t - P^{K^\ast}_t) (A_t - B_t K_t) + K^\dagger_t R_t K_t \\
        & - \big[B_t(K_t - K^\ast_t)\big]^\dagger P^{K^\ast}_t - P^{K^\ast}_t \big[B_t(K_t - K^\ast_t)\big] - (K^\ast_t)^\dagger R_t K^\ast_t,
    \end{align*}
    with the terminal condition $P^K_T - P^{K^\ast}_T = 0$.  Integrating over $[t, T]$ and taking norms on both sides:
    \begin{multline*}
        ||P^K_t - P^{K^\ast}_t||_2 \leq 2 \int_t^T ||A_s - B_s K_s||_2 ||P^K_s - P^{K^\ast}_s||_2 \dd s + 2 \max_{s \in [0, T]} ||B_s^\dagger P^{K^\ast}_s||_2 \int_t^T ||K_s - K^\ast_s||_2 \dd s \\
        + \max_{s \in [0, T]} ||R_s||_2 \int_t^T \big(||K_s||_2 + ||K^\ast_s||_2\big) ||K_s - K^\ast_s||_2 \dd s.
    \end{multline*}
    Recall from the proof of Lemma~\ref{lemma: lower bound} that $\big|\big|K_s\big|\big|_2 \leq \big|\big|R^{-1}_s B^\dagger_s\big|\big|_2 \big|\big|P^{K^{(0)}}_s\big|\big|_2$. Therefore,
    \begin{align*}
        ||P^K_t - P^{K^\ast}_t||_2 & \leq 2 \int_t^T ||A_s - B_s K_s||_2 ||P^K_s - P^{K^\ast}_s||_2 \dd s + \bigg[ 2 \max_{s \in [0, T]} ||B_s^\dagger P^{K^\ast}_s||_2 \\
        & + \max_{s \in [0, T]} ||R_s||_2 \Big(\max_{s \in [0, T]} \big|\big|R^{-1}_s B^\dagger_s\big|\big|_2 \big|\big|P^{K^{(0)}}_s\big|\big|_2 + \max_{s \in [0, T]} ||K^\ast_s||_2 \Big)\bigg] \int_t^T ||K_s - K^\ast_s||_2 \dd s.
    \end{align*}
    Therefore, by Gronwall's inequality, there exists $\bar{c}_\rho > 0$, which is affected by $K^{(0)}$, such that 
    \[
    \max_{t \in [0, T]} ||P^K_t - P^{K^\ast}_t||_2 \leq \bar{c}_\rho \int_0^T ||K_t - K^\ast_t||_2 \dd t,
    \]
    and moreover,
    \[
    \max_{t \in [0, T]}||K^\prime_t - K^\ast_t||_2  \leq \bar{c}_\rho \max_{t \in [0, T]} \frac{||B_t||_2} {\lambda_{\min}(R_t)} \int_0^T ||K_t - K^\ast_t||_2 \dd t.
    \]
    Finally, noticing that $\int_0^T ||A_t - B_t K^\ast_t||_2 \dd t \leq \rho$, an application of Lemma~\ref{lemma: super 2} finishes the proof.
\end{proof}

\begin{proof}[Proof of Proposition~\ref{prop: IPO_local}]
    To show the existence of $\epsilon$, denote $r := \int_0^T ||K^{(0)}_t - K^\ast_t||_2^2 \dd t$. Recall from the proof of Lemma~\ref{lemma: lower bound} that $||K^{(i)}_t||_2 \leq ||R^{-1}_t B^\dagger_t||_2 ||P^{K^{(0)}}_t||_2$ for any $i \geq 1$. By applying Gronwall's inequality on \eqref{eq: P_t}, $\max_{t \in [0, T]} ||P^{K^{(0)}}_t||_2$ is bounded from above, and the bound only depends on the value of $r$ (as an increasing function in $r$) and the data of the LQR. As a result, there exists $\rho_r > 0$, which only depends on $r$ (as an increasing function in $r$) and the data of the LQR, such that
    \[
    \max\left\{\int_0^T ||A_t - B_t K^\ast_t||_2 \dd t,\, \sup_{i \geq 0} \int_0^T ||A_t - B_t K^{(i)}_t||_2 \dd t\right\} \leq \rho_r.
    \]

    By Lemma~\ref{lemma: lower bound} and Lemma~\ref{lemma: IPO_K_diff}, there exists $\underline{\mu}_r > 0$, which only depends on the value of $r$ (as a decreasing function in $r$) and the data of the LQR, such that
    \[
    \forall i \geq 0, \quad \underline{\mu}_r \delta \int_0^T ||K^{(i)}_t - K^\ast_t ||_2^2 \dd t \leq C(K^{(i)}, \Sigma) - C(K^\ast, \Sigma).
    \]
    Meanwhile, by applying Gronwall's inequality on \eqref{eq: y_t_ode}, there exists $\bar{\mu}_r > 0$, which only depends on the value of $r$ (as an increasing function in $r$) and the data of the LQR, such that
    \[
    \forall i \geq 0, \quad \max_{t \in [0, T]} ||y^{(i)}_t||_2 \leq \bar{\mu}_r. 
    \]
    As a result, by Lemma~\ref{lemma: cost difference}, 
    \[
    \forall i \geq 0, \quad C(K^{(i)}, \Sigma) - C(K^\ast, \Sigma) \leq \bar{\mu}_r \max_{t \in [0, T]} ||R_t||_2 \int_0^T ||K^{(i)}_t - K^\ast_t ||_2^2 \dd t. 
    \]
    By Lemma~\ref{lemma: super 3}, there exists $c^\ast_r > 0$, which only depends on the value of $r$ (as an increasing function in $r$) and the data of the LQR, such that 
    $$
    \forall i \geq 0, \quad \max_{t \in [0, T]} ||y^{(i + 1)}_t - y^\ast_t||_2 \leq c^\ast_r \int_0^T ||K^{(i)}_t - K^\ast_t||_2 \dd t.
    $$
    Therefore, for any $i \geq 0$, we have
    \begin{align*}
        \max_{t \in [0, T]} ||y^{(i + 1)}_t - y^\ast_t||_2 & \leq c^\ast_r \sqrt{T} \sqrt{\int_0^T ||K^{(i)}_t - K^\ast_t||^2_2 \dd t}  \\
        & \leq c^\ast_r\sqrt{\frac{T}{\underline{\mu}_r \delta}} \sqrt{C(K^{(0)}, \Sigma) - C(K^\ast, \Sigma)} \\
        & \leq c^\ast_r\sqrt{\frac{T\bar{\mu}_r \max_{t \in [0, T]} ||R_t||_2}{\underline{\mu}_r \delta}} \sqrt{r}. 
    \end{align*}
    Since the RHS is an increasing function in $r$ and tends to 0 as $r \rightarrow 0^+$, there exists $\epsilon > 0$, such that $r < \epsilon$ implies 
    \[
    \forall i \geq 1, \quad \max_{t \in  [0, T]}||y^{(i)}_t - y^\ast_t||_2 \leq \min_{t \in [0, T]} \lambda_{\min}(y^\ast_t).
    \]
    This proves the existence of $\epsilon$.
    
    Finally, to calculate the corresponding $\mathcal{C}_2$, by Lemma~\ref{lemma: super 1}, 
    \begin{align}
        \forall i \geq 0, \quad C (K^{(i + 1)}, & \Sigma) - C(K^\ast, \Sigma) \nonumber \\
        & \leq \frac{\max_{t \in  [0, T]} ||y^{(i + 1)}_t - y^\ast_t ||_2}{\min_{t \in [0, T]}\lambda_{\min}(y^\ast_t)} \left[C(K^{(i)}, \Sigma) - C(K^\ast, \Sigma)\right] \nonumber \\
        & \leq \frac{c^\ast_r \sqrt{T}}{\min_{t \in [0, T]}\lambda_{\min}(y^\ast_t)} \sqrt{\int_0^T ||K^{(i)}_t - K^\ast_t||^2_2 \dd t} \left[C(K^{(i)}, \Sigma) - C(K^\ast, \Sigma)\right] \nonumber \\
        & \leq \frac{c^\ast_r}{\min_{t \in [0, T]}\lambda_{\min}(y^\ast_t)} \sqrt{\frac{T}{\underline{\mu}_r \delta}} \left[C(K^{(i)}, \Sigma) - C(K^\ast, \Sigma)\right]^{\frac{3}{2}}, \nonumber
    \end{align}
    \ie $\mathcal{C}_2 = \frac{c^\ast_r }{\min_{t \in [0, T]}\lambda_{\min}(y^\ast_t)} \sqrt{\frac{T}{\underline{\mu}_r \delta}}$. 
\end{proof}

\bibliography{refs}
\bibliographystyle{unsrt}

\appendix
\section{Preliminaries of the rough path theory} \label{app: prelim}

In this section, we review some results in the \textit{rough path theory} which are relevant to our proof of Proposition~\ref{prop: existence} and Theorem~\ref{thm: general_transfer} in Section~\ref{sec: general}. The main reference of this section is \cite{friz2010multidimensional}, and we refer to \cite{kirillov2008introduction} for an introduction of Lie groups and Lie algebras.

\section{Truncated tensor algebra and its subspaces}

Let $d \geq 1$ be the dimension and $N \geq 0$ be the level of truncation. We define
\[
T^N(\R^d) := \bigoplus_{k=0}^N (\R^d)^{\otimes k}
\]
as the space of tensor algebra truncated at level $N$. It is not hard to verify that $(T^N(\R^d), +, \cdot; \otimes)$ is an associative algebra (\cf \cite[Proposition~7.4]{friz2010multidimensional}). In the following, we shall use $\pi_k: T^N(\R^d) \to (\R^d)^{\otimes k}$ to denote the $k$-th level projection map ($0 \leq k \leq N$).

Now, define the following subspace of $T^N(\R^d)$: 
\[
1 + \mathfrak{t}^N(\R^d) := \big\{ g \in T^N(\R^d) : \pi_0(g) = 1 \big\}.
\]
Note that $1 + \mathfrak{t}^N(\R^d)$ is a Lie group with the group multiplication $\otimes$ and the manifold topology is induced by the uniform norm as a finite-dimensional vector space (\cf \cite[Proposition~7.17]{friz2010multidimensional}). Its Lie algebra is explicitly written out by
\[
\mathfrak{t}^N(\R^d) := \big\{ g \in T^N(\R^d) : \pi_0(g) = 0 \big\},
\]
where the Lie bracket $[\cdot, \cdot]: \mathfrak{t}^N(\R^d) \times \mathfrak{t}^N(\R^d) \to \mathfrak{t}^N(\R^d)$ is given by the commutator, \ie
\[
\forall x, y \in \mathfrak{t}^N(\R^d), \quad [x, y] := x \otimes y - y \otimes x,
\]
and the exponential map $\exp: \mathfrak{t}^N(\R^d) \to 1 + \mathfrak{t}^N(\R^d)$ is given by
\[
\forall x \in \mathfrak{t}^N(\R^d), \quad \exp(x) := 1 + \sum_{k=1}^N \frac{x^{\otimes k}}{k!}.
\]
In this case, the logarithm map $\log: 1 + \mathfrak{t}^N(\R^d) \to \mathfrak{t}^N(\R^d)$ is globally defined and writes
\[
\forall g \in 1 + \mathfrak{t}^N(\R^d), \quad \log(g) := \sum_{k=1}^N (-1)^{k+1} \frac{(g - 1)^{\otimes k}}{k}.
\]

Next, define $\mathfrak{g}^N(\R^d) \subseteq \mathfrak{t}^N(\R^d)$ by the smallest Lie subalgebra that contains $\pi_1(\mathfrak{t}^N(\R^d))$, \ie
\[
\mathfrak{g}^N(\R^d) := \R^d \oplus [\R^d, \R^d] \oplus \cdots \oplus [\R^d, [\cdots, [\R^d, \R^d]]],
\]
where the rightmost term contains $(N - 1)$ Lie brackets. Define
\[
G^N(\R^d) := \exp(\mathfrak{g}^N(\R^d)).
\]
Then, by \cite[Theorem~7.30]{friz2010multidimensional}, $G^N(\R^d)$ is a closed Lie subgroup of $1 + \mathfrak{t}^N(\R^d)$, equipped with the submanifold topology. Let $||\cdot||$ denote the \textit{Carnot-Caratheodory norm} on $G^N(\R^d)$, \ie
\[
\forall g \in G^N(\R^d), \quad ||g|| := \inf\left\{\int_0^1 |\dd\gamma_t| : \gamma \in C^{1\text{-var}}([0, 1], \R^d) \text{ such that } S_N(\gamma_{[0, 1]}) = g\right\},
\]
where $C^{1\text{-var}}([0, 1], \R^d)$ denotes the space of bounded variation continuous paths (recall that $\int_0^1 |\dd\gamma_t|$ is the total variation of $\gamma$), and $S_N(\gamma_{[0, 1]})$ denotes the truncated \textit{signature} of $\gamma$:
\begin{align} \label{eq: signature}
    S_N(\gamma_{[0, 1]}) := \bigg(1, \int_0^1 \dd \gamma_t, & \int_0^1 \int_0^{t_2} \dd \gamma_{t_1} \otimes \dd\gamma_{t_2}, \cdots, \nonumber \\
    & \int_0^1 \int_0^{t_N} \cdots \int_0^{t_2} \dd \gamma_{t_1} \otimes \cdots \otimes \dd \gamma_{t_N} \bigg) \in G^N(\R^d).
\end{align}
By \cite[Proposition~7.42]{friz2010multidimensional}, $||\cdot||$ makes $G^N(\R^d)$ a geodesic space. Note that $||\cdot||$ induces the same topology on $G^N(\R^d)$ as the submanifold topology (\cf \cite[Corollary~7.46]{friz2010multidimensional}).

\section{Geometric \texorpdfstring{$p$}{}-rough paths}

Let $p \geq 1$ and $[p]$ denotes the integer part of $p$. Recall that $G^{[p]}(\R^d)$ is a Lie group and $||\cdot||$ denotes the Carnot-Caratheodory norm on it. The space of \textit{weak geometric $p$-rough paths} is defined by
\[
C^{p\text{-var}} ([0, T], G^{[p]}(\R^d)) := \Big\{ \gamma \in C([0, T], G^{[p]}(\R^d)) : ||\gamma||_{p\text{-var}; [0, T]} < +\infty\Big\},
\]
where
\[
||\gamma||_{p\text{-var}; [0, T]} := \sup_{\{t_i\}_{i=0}^n \subset [0, T]} \left( \sum_{i=0}^{n-1} ||\gamma_{t_i}^{-1} \otimes \gamma_{t_{i+1}}||^p\right)^{\frac{1}{p}}.
\]
We note that $||\cdot||_{p\text{-var}; [0, T]}$ is only a seminorm since all constant paths have norm 0. Next, we equip $C^{p\text{-var}} ([0, T], G^{[p]}(\R^d))$ with the true metric $d_{p\text{-var}; [0, T]}$, given by
\begin{align} \label{eq: norm}
    & \forall \gamma, \tilde\gamma \in C^{p\text{-var}} ([0, T], G^{[p]}(\R^d)), \nonumber \\
    & d_{p\text{-var}; [0, T]}(\gamma, \tilde\gamma) := ||\gamma_0^{-1} \otimes \tilde\gamma_0|| + \sup_{\{t_i\}_{i=0}^n \subset [0, T]} \left(\sum_{i=0}^{n-1} ||(\gamma_{t_i}^{-1} \otimes \gamma_{t_{i+1}})^{-1} \otimes (\tilde\gamma_{t_i}^{-1} \otimes \tilde\gamma_{t_{i+1}})||^p \right)^{\frac{1}{p}},  
\end{align}
where we use the subscript $_t$ to refer to the time $t$ point of the corresponding curve. It is shown that $C^{p\text{-var}} ([0, T], G^{[p]}(\R^d))$ is a complete non-separable metric space (\cf \cite[Theorem~8.13(i)]{friz2010multidimensional}). 

With an abuse of notation, for a given smooth path $\gamma \in C^\infty([0, T], \R^d)$, we use $S_{[p]}(\gamma)$ to denote the corresponding lifted path on $G^{[p]}(\R^d)$, \ie
\[
S_{[p]}(\gamma)_t := S_{[p]}(\gamma_{[0, t]}), \quad t \in [0, T],
\]
where the truncated signature transform on the RHS is defined in \eqref{eq: signature}. Next, the space of \textit{geometric $p$-rough paths}, denoted by $C^{0, p\text{-var}} ([0, T], G^{[p]}(\R^d))$, is defined by the closure of (the lift of) smooth paths under the metric \eqref{eq: norm}. To spell out, $\gamma \in C^{0, p\text{-var}} ([0, T], G^{[p]}(\R^d))$ if and only if there exists a sequence $\{\gamma^{(i)}\}_{i=0}^\infty$ of smooth paths in $C^\infty([0, T], \R^d)$ such that
\[
\lim_{i \to \infty} d_{p\text{-var}; [0, T]}\big(\gamma^{-1}_0 \otimes \gamma, S_{[p]}(\gamma^{(i)})\big) = 0.
\]
It is shown that $C^{0, p\text{-var}} ([0, T], G^{[p]}(\R^d)) \subset C^{p\text{-var}} ([0, T], G^{[p]}(\R^d))$, \ie the inclusion is strict (\cf \cite[Corollary~8.24]{friz2010multidimensional}), and $C^{0, p\text{-var}} ([0, T], G^{[p]}(\R^d))$ is a complete separable metric (\ie Polish) space (\cf \cite[Proposition~8.25]{friz2010multidimensional}). Finally, we mention that when $p > 1$, $C^{0, p\text{-var}} ([0, T], G^{[p]}(\R^d))$ can be equivalently defined by the closure of (the lift of) $C^{1\text{-var}}([0, T], \R^d)$ paths under $d_{p\text{-var}; [0, T]}$ (\cite[Lemma~8.21]{friz2010multidimensional}).

\begin{example}[Enhanced Brownian motion] \label{exam: BM}
    Let $W_t$ be a $d$-dimensional BM. Then, the enhanced BM
    \[
    \mathbf{W}_t := \left(1, W_t, \int_0^t W_s \otimes \circ \dd W_s\right)
    \]
    is a geometric $p$-rough path for any $2 < p < 3$ almost surely (\cf \cite[Section~13.2]{friz2010multidimensional}), where $\circ \dd W_s$ stands for the Stratonovich integral.
\end{example}

\section{Rough differential equations (RDEs)}

We restate the definition of \textit{rough differential equations} (RDEs) in \cite[Definition~10.17]{friz2010multidimensional} below.

\begin{definition}[Rough differential equations]
    Let $p \geq 1$. Suppose $\mathbf{x} \in C^{p\text{-var}} ([0, T], G^{[p]}(\R^d))$ is a weak geometric $p$-rough path and $V: \R^e \to (\R^d)^\ast$ is a given vector field. Then, we say that $y \in C([0, T], \R^e)$ is a solution to the RDE
    \begin{equation} \label{eq: RDE_y}
        \dd y_t = V(y_t) \dd \mathbf{x}_t, \quad y_0 \in \R^e \text{ given}
    \end{equation}
    if there exists a sequence $\{x^{(i)}\}_{i=0}^\infty$ of paths in $C^{1\text{-var}}([0, T], \R^d)$ and a sequence $\{y^{(i)}\}_{i=0}^\infty$ of paths in $C([0, T], \R^e)$ such that 
    \begin{enumerate}[1)]
        \item $\lim_{i \to \infty} d_{0; [0, T]} (S_{[p]}(x^{(i)}), \mathbf{x}) = 0 \text{ and } \sup_{i} ||S_{[p]}(x^{(i)})||_{p\text{-var}; [0, T]} < +\infty$;
        \item $y^{(i)}_0 = y_0$ and $\dd y^{(i)}_t = V(y^{(i)}_t) \dd x^{(i)}_t$ for any $i \geq 0$;
        \item $y^{(i)} \rightarrow y \text{ uniformly on $[0, T]$ as $i \rightarrow \infty$}$,
    \end{enumerate}
    where 
    \begin{equation*}
        \forall \mathbf{y}, \mathbf{z} \in C ([0, T], G^{[p]}(\R^d)), \quad d_{0; [0, T]} (\mathbf{y}, \mathbf{z}) := \sup_{0 \leq s < t \leq T} ||(\mathbf{y}_s^{-1} \otimes \mathbf{y}_t)^{-1} \otimes (\mathbf{z}_s^{-1} \otimes \mathbf{z}_t)||.
    \end{equation*}
\end{definition}

Finally, we refer to \cite[Sections~10--12]{friz2010multidimensional} for more details of relevant theories on the RDE \eqref{eq: RDE_y} (\eg, its existence, uniqueness, and stability). 

\end{document}

%% file: defs.tex
\newcommand{\BEAS}{\begin{eqnarray*}}
\newcommand{\EEAS}{\end{eqnarray*}}
\newcommand{\BEQ}{\begin{equation}}
\newcommand{\EEQ}{\end{equation}}
\newcommand{\BIT}{\begin{itemize}}
\newcommand{\EIT}{\end{itemize}}

\newcommand{\eg}{\emph{e.g.}, }
\newcommand{\ie}{\emph{i.e.}, }

\newcommand{\cf}{\emph{cf.} }
\newcommand{\resp}{\emph{resp.} }

\newcommand{\E}{\mathbb{E}}

\newcommand{\argmin}{\mathop{\rm argmin}}

\def\<#1,#2>{\langle #1,#2\rangle}

\newtheorem{theorem}{Theorem}
\newtheorem{remark}{Remark}
\newtheorem{assumption}{Assumption}
\newtheorem{lemma}[theorem]{Lemma}
\newtheorem{corollary}[theorem]{Corollary}
\newtheorem{proposition}[theorem]{Proposition}
\newtheorem{definition}{Definition}
\newtheorem{example}{Example}

{\begin{quote}}{\end{quote}}

\makeatletter
\long\def\@makecaption#1#2{
   \vskip 9pt
   \begin{small}
   \setbox\@tempboxa\hbox{{\bf #1:} #2}
   \ifdim \wd\@tempboxa > 5.5in
        \begin{center}
        \begin{minipage}[t]{5.5in}
        \addtolength{\baselineskip}{-0.95pt}
        {\bf #1:} #2 \par
        \addtolength{\baselineskip}{0.95pt}
        \end{minipage}
        \end{center}
   \else
	\hbox to\hsize{\hfil\box\@tempboxa\hfil}
   \fi
   \end{small}\par
}
\makeatother

\newcounter{oursection}

\newcounter{lecture}

\newcommand{\A}{\mathcal{A}}

\newcommand{\dd}{\mathrm{d}}
\newcommand{\D}{\mathcal{D}}
\newcommand{\F}{\mathcal{F}}
\newcommand{\G}{\mathcal{G}}

\newcommand{\PP}{\mathbb{P}}
\newcommand{\pP}{\mathcal{P}}
\newcommand{\R}{\mathbb{R}}

\newcommand{\pd}[2]{\frac{\partial #1}{\partial #2}}

\DeclareMathOperator{\tr}{tr}